\documentclass[3p,article]{elsarticle}
\usepackage[utf8]{inputenc}
\usepackage{algorithm}
\usepackage{algorithmic}

\usepackage{hyperref}
\usepackage{times}
\usepackage{amssymb}
\usepackage{array,color}
\usepackage{booktabs}
\usepackage{multirow}
\usepackage{chngpage}
\usepackage{lscape}
\usepackage{amsfonts}
\usepackage{amsmath}
\usepackage{enumitem}
\usepackage{graphicx}
\usepackage{bm}
\usepackage[caption=false,font=footnotesize]{subfig}
\usepackage{float}
\biboptions{numbers,sort&compress}

\usepackage{tabularx}      
\usepackage{adjustbox}     
\usepackage{tablefootnote} 

\newcounter{promptbox}
\usepackage{caption}          

\usepackage{listings}

\usepackage[most]{tcolorbox}  
\definecolor{boxblue}{RGB}{0,126,214}
\definecolor{boxbg}{RGB}{235,245,255}

\lstdefinestyle{prompttiny}{
  basicstyle=\scriptsize\ttfamily\selectfont,  
  breaklines=true,
  breakatwhitespace=true,
  keepspaces=true,
  columns=fullflexible
}

\newtcolorbox{promptmini}[2][]{
  enhanced, 
  colback=boxbg, 
  colframe=boxblue,
  coltitle=white, 
  colbacktitle=boxblue,
  fonttitle=\bfseries\footnotesize,
  attach boxed title to top left={xshift=4mm,yshift*=-1mm},
  arc=1mm, boxed title style={arc=1mm},
  left=2mm, right=2mm, top=1.5mm, bottom=1.5mm,
  boxrule=0.6pt,
  listing only,                         
  listing options={style=prompttiny},   
  width=\columnwidth,
    before upper = {\refstepcounter{promptbox}},
  #1}

\begin{document}

\begin{frontmatter}

\title{
Offline Multi-Task Multi-Objective Data-Driven Evolutionary Algorithm with Language Surrogate Model and Implicit Q-Learning
}
\tnotetext[mytitlenote]{
    This work was supported in part by National Natural Science Foundation of China (Grant No.62276100), in part by Guangzhou Science and Technology Elite Talent Leading Program for Basic and Applied Basic Research (Grant No. SL2024A04J01361), in part by the Guangdong Provincial Natural Science Foundation for Outstanding Youth Team Project (Grant No. 2024B1515040010), in part by the Fundamental Research Funds for the Central Universities (Grant No. 2025ZYGXZR027) and in part by the National Research Foundation of Korea (Grant No. NRF2022H1D3A2A01093478).
}
\author[aff1]{Xian-Rong Zhang}
\author[aff1]{Yue-Jiao Gong}
\author[aff1]{Zeyuan Ma\corref{mycorrespondingauthor}}
\cortext[mycorrespondingauthor]{Corresponding author (scut.crazynicolas@gmail.com)}
\author[aff2,aff3]{Jun Zhang}

\address[aff1]{School of Computer Science and Engineering, South China University of Technology, Guangzhou, China.}
\address[aff2]{College of Artificial Intelligence, Nankai University, Tianjin, China.}
\address[aff3]{Hanyang University, 15588 Ansan, South Korea.}

\begin{abstract}
Data-driven evolutionary algorithms has shown surprising results in addressing expensive optimization problems through robust surrogate modeling. Though promising, existing surrogate modeling schemes may encounter limitations in complex optimization problems with many sub-objectives, which rely on repeated and tedious approximation. To address such technical gap, we propose Q-MetaSur as a plug-and-play surrogate modeling scheme capable of providing unified and generalized surrogate learning. Specifically, we consider multi-task-multi-objective optimization~(MTMOO) in offline setting. Several key designs are proposed: 1) we transform objective approximation into sequence-to-sequence modeling where MTMOO problem can be represented by tenxual tokenization. To operate under such auto-regressive modeling, we introduce a Large Language Model-based surrogate model that first encodes a MTMOO instance and then decodes objective values of unseen decision variables. To ensure stability in training the proposed model, we propose a two-stage offline training strategy that operates as a synergy of supervised tuning and RL fine-tuning, which first exploits offline dataset to fit existing knowledge and then leverages RL to enhance model's generalization performance. Extensive empirical results on the CEC2019 benchmark demonstrate that Q-MetaSur not only outperforms representative surrogate baselines in objective approximation accuracy, but also helps underlying evolutionary algorithms achieve both desired optimization convergence and improved pareto optimality.
\end{abstract}


\end{frontmatter}


\section{Introduction}
Many real-world expensive multi-objective optimization problems (MOPs) involve conflicting objectives and often lack explicit closed-form expressions for their objective and constraint functions. In such cases, optimization must rely on limited data obtained from physical experiments, real-world operations, or computationally intensive numerical simulations. For example, one evaluation of a high-fidelity vehicle crashworthiness analysis can require several days of computation, so performing $10^4$ crash simulations for a single design study would take more than 100 years~\cite{DDEA-PES}. A common strategy is to construct surrogate models from these historical data and embed them into multi-objective evolutionary algorithms (MOEAs), so that the surrogates can replace costly ground-truth evaluations during the search. This forms the basic paradigm of data-driven evolutionary algorithms (DDEAs)~\cite{DDEA_overview,DDEA}. In online DDEAs, additional (and expensive) evaluations can be performed during optimization to update the surrogate, whereas offline DDEAs rely solely on a fixed dataset collected \emph{a priori}. This paper focuses on the more challenging offline setting, where the generalization ability of the surrogate directly determines the quality of the resulting Pareto front. A variety of offline data-driven multi-objective evolutionary algorithms (DD-MOEAs, i.e., offline multi-objective DDEAs) have been developed for engineering data-driven multi-objective optimization problems (DD-MOPs), including trauma system design, process industry optimization, and hardware accelerator design~\cite{trauma_systems,fused_magnesium_furnaces,blast__furnace,Ceramic_Formula,Hardware_Accelerators}.

To better exploit cross-task structural information under a limited evaluation budget, evolutionary multi-task optimization has attracted increasing attention in recent years. Its goal is to solve multiple MOPs within a unified framework and obtain a set of Pareto-optimal solutions across tasks.
Gupta \emph{et al.}~\cite{MO-MFEA} introduced multifactorial evolutionary optimization for multi-task multi-objective optimization, and subsequent works such as ExTrEMO~\cite{ExTrEMO} further targeted expensive multi-task multi-objective optimization (MTMOO) scenarios.
Because of the potential similarities between tasks, multi-objective multi-task optimization typically achieves more competitive solution sets than treating each DD-MOP as an isolated ``one-shot'' task under the same evaluation budget. This naturally motivates a corresponding need for \emph{joint multi-task and multi-objective modeling at the surrogate level}.
From the surrogate modeling perspective, multi-task Gaussian processes (MTGPs) are a representative approach for multi-objective multi-task settings~\cite{multi_output_Gaussian_process,NIPS2007_MTGP,NIPS2008_MTGP_CNN,MTGP_CNNMix_TNNLS}. MTGPs explicitly model inter-task correlations via structured covariance functions, and have been successfully applied to Bayesian optimization and multi-task optimization~\cite{MTGP_Bayesian,MTGP_Bayesian2,MTGP_Bayesian3,Heter_Input_MTGP,Multiview_GP,Heter-TGP}. However, in expensive and offline multi-task multi-objective scenarios, classical MTGPs face pronounced scalability bottlenecks: for $T$ tasks and $TN$ samples, the time complexity is $\mathcal{O}(T^3 N^3)$, and remains formidable even with approximate methods~\cite{sparse_MTGP,large_MTGP,pmlr-v130-wang21c}. In addition, MTGPs struggle to handle tasks with highly heterogeneous input dimensions and feature representations, which require complex cross-domain alignment and are prone to negative transfer~\cite{Heter_Input_MTGP,Multiview_GP,Heter-TGP}. More importantly, most existing methods focus on modeling correlations in the \emph{task} dimension, while still relying on separate models or outputs for different objectives in the \emph{objective} dimension, thus failing to fully exploit the coupled information across multiple objectives.

In this paper, we propose Q-MetaSur, an LLM-based meta-surrogate model tailored for offline multi-task multi-objective optimization. We encode task metadata, decision variables, and multi-objective fitness values into a single textual sequence, where all numerical values are represented in scientific notation and different objectives occupy explicitly delineated segments in the output sequence. This design allows a single LLM to jointly model multiple tasks and a variable number of objectives without modifying the backbone architecture. On top of this representation, we adopt a two-stage fine-tuning paradigm. First, we perform supervised fine-tuning (SFT) with priority-weighted cross-entropy to learn a stable ``numerical language.'' Then, in the reinforcement learning fine-tuning (RLFT) stage, we incorporate Conservative Q-Learning (CQL) in an offline setting to directly optimize a sequence-level reward defined by range-normalized multi-objective errors, thereby aligning the training objective more closely with the desired continuous numerical accuracy. When integrated with multi-objective multi-task evolutionary algorithms, the resulting framework achieves knowledge transfer at both the surrogate level and the individual level, and is expected to improve the quality and stability of the Pareto front under a fixed evaluation budget.

Against this background, the main contributions of this work can be summarized as follows:
\begin{itemize}
  \item \textbf{A unified LLM-based meta-surrogate framework for multi-task multi-objective modeling.}
  We propose a unified textual representation in scientific notation that encodes task metadata, decision variables, and a variable number of objective values into a single token sequence. This enables one LLM to share knowledge across both the task and objective dimensions, alleviating the information fragmentation caused by per-objective surrogate models in existing approaches.
  \item \textbf{Two-Stage Offline Training Strategy.}
  To ensure training robustness and stable convergence, we further introduce a two-stage offline training strategy that synergistically couples supervised fine-tuning with Conservative Q-Learning (CQL). In the first stage, the model leverages the available offline dataset to thoroughly fit the observed knowledge. In the second stage, CQL is employed to enhance the model’s generalization ability under regularized distributional shift.
  \item \textbf{Perturb-and-score augmentation and emergent generalization.}
  Around each ground-truth label vector, we synthesize and score candidate sequences, expanding a limited offline dataset into a rich set of (sequence, reward) pairs. This improves the numerical quality of the Pareto front constructed from the surrogate without additional ground-truth evaluations. At the same time, Q-MetaSur exhibits meaningful zero-/few-shot generalization on unseen tasks and new decision-space dimensions.
  \item \textbf{End-to-end integration and evaluation for multi-task multi-objective optimization.}
  We seamlessly integrate Q-MetaSur with existing MTMOO algorithms to construct an offline data-driven multi-task multi-objective optimization framework, and systematically evaluate its surrogate accuracy and optimization performance on multi-task multi-objective benchmarks.
\end{itemize}

The remainder of this paper is organized as follows. \autoref{sec:bg} reviews the research background and related work. \autoref{sec:method} presents the proposed model and training paradigm. \autoref{sec:experiments} describes the experimental setup and results on benchmark multi-task multi-objective problems. In addition, \autoref{sec:app_experiments} presents an application study on sensor coverage optimization. Finally, \autoref{sec:conclusion} concludes the paper and discusses future research directions.

\section{Background and Related Work}
\label{sec:bg}

\subsection{Data-Driven Multi-Objective Optimization}
Many real-world optimization problems encompass more than one objective functions to be optimized. These problems, commonly referred to as multi-objective optimization (MOO) problems, may hold entirely conflict objectives, posing significant challenges for global optimization techniques such s evolutionary algorithms.  A MOO problem can be formulated as:
\begin{equation}
    F(x) = ( f^{(1)}(x),\, \ldots,\, f^{(j)}(x),\, \ldots,\, f^{(k)}(x) ).
\end{equation}
Here, $x \in \mathbb{R}^{n}$ is an $n$-dimensional decision vector, $k$ denotes the number of objectives, $f^{(j)}(x)$ is the $j$-th objective function, and $F(x)$ is the objective vector of the entire optimization problem. To address MOO problems, evolutionary algorithms MOEA/D~\cite{MOEA/D-MTGP} and NSGA-II~\cite{NSGA-II}, and their modern variants~\cite{REAV} have been proposed and applied in both scientific discovery~\cite{Chemical} and industrial production~\cite{DDMOEA-GAN-offline} scenarios. 

However, due to the complexity of physical or chemical processes in practical engineering applications, the objective functions in these problems are often difficult to model explicitly. That is, when time or cost is prohibitive, only a small number of samples can be obtained through expensive experiments~(either historical data or online evaluation). In literature, such expensive-evaluation case is often addressed by Data-Driven Evolutionary Algorithm~(DDEA)~\cite{DDEA}. In DDEA, surrogate model is introduced to approximate the objective functions, and then the well-trained surrogate models could be used to substitute original multiple objectives in a MOO problem. 
In specific, given a collection of $N$ evaluation data $\mathcal{D} := \{(x_i,F(x_i))\}_{i=1}^{N}$ and a group of learnable surrogate models with parameters denoted as $\theta$, the evaluation in DDEA can be formulated as:
\begin{equation}
    \hat{F}(x\mid\mathcal{D},\theta) = ( \hat{f}^{(1)}(x),\, \ldots,\, \hat{f}^{(j)}(x),\, \ldots,\, \hat{f}^{(k)}(x) ),
\end{equation}
which is also a $k$-dimensional objective vector.
A naive way to measure the accuracy of the approximated objective vector is computing the euclidean distance between $F(\cdot)$ and $\hat{F}(\cdot)$. However, since the use of surrogate models in DDEA may directly impacts the evolutionary optimization process, existing literature leans to use the performance difference~(convergence and diversity) between optimization with $F(\cdot)$ and with $\hat{F}(\cdot)$. Specifically, in DDEA approach for MOO problems, the accuracy of approximating $F(x)$ by $\hat{F}(x)$ can be defined analogously to the inverted generational distance (IGD)~\cite{IGD}, where pareto set induced by $\hat{F}(x)$ and $F(x)$ correspond to an approximated Pareto front (PF) and the true PF, respectively. We denote the former pareto set as $A$ and the latter one as $T$, then the accuracy of $\hat{F}(x)$ in approximating $F(x)$ can be computed as: 
\begin{equation}
    \mathrm{IGD}(T, A) = \frac{1}{|T|} \sum_{i=1}^{|T|} \min_{1 \le j \le |A|} d(t_i, a_j),
\end{equation}
Here, $A = \{a_1, a_2, \ldots, a_{|A|}\}$ denotes the approximate PF obtained from $\hat{F}(x)$, where $a_j$ is the $j$-th pareto optimal; $T = \{t_1, t_2, \ldots, t_{|T|}\}$ denotes the true PF obtained from $F(x)$, where $t_i$ is the $i$-th pareto optimal; $d(t_i, a_j)$ denotes the euclidean distance between two points in the objective space. The primary goal is to construct a high-quality $\hat{F}(x)$ such that minimizes the accuracy indicator, which in turn impacts final optimization performance.

Following such surrogate-assisted principle, existing DDEA approaches tailored for MOO problems can be categorized into two lines depending on whether new data can be acquired via true fitness evaluations during the optimization process. One category allows sampling a certain amount of new data through expensive fitness evaluations during evolutionary optimization, known as online DDEA~\cite{online_ddea}.
In such cases, traditional MOEAs are often inapplicable due to their reliance on numerous true fitness evaluations~\cite{MOEA_multiFE}. Evolutionary algorithms that reduce sample consumption by approximating fitness evaluations with surrogate predictors are called surrogate-assisted evolutionary algorithms (SAEAs). Many successful SAEAs for expensive multi-objective optimization have been proposed, such as SPGP~\cite{SPGP}, HES-EA~\cite{HES-EA}, and EMMOEA~\cite{EMMOEA}, among others.
However, in some scenarios, additional active experimentation or simulation (i.e., true fitness evaluation) is not allowed during optimization; this is called offline DDEA~\cite{DDMOEA-GAN-offline}. The challenge lies in the fact that only a small amount of historical offline data is available in this setting, which makes online DDEA inapplicable. For example, in fused magnesium furnace optimization~\cite{fused_magnesium_furnaces}, the target value of power consumption per ton of magnesite cannot be freely obtained due to complex environmental simulation: 1) at high temperatures, mixtures of solid, liquid, and gas phases may cause various sensor failure; 2) magnesite production involves multiple physical and chemical processes, such as melting, impurity precipitation, and crystallization, which results in impractical online evaluation. Representative offline multi-objective DDEA studies include DDMOEA-GAN~\cite{DDMOEA-GAN-offline}, TGPR-MO~\cite{TGPR-MO-offline}, and IBEA-MS~\cite{IBEA-MS-offline}, among others.

\subsection{Multi-Task Multi-Objective Optimization}
Though existing offline multi-objective DDEAs present certain performance advantage in their own target problems, they are avoidably restricted by modern complex optimization environment, where many optimization tasks anticipate quick solving simultaneously. This multi-task setting was first defined by Gupta. et al. in ~\cite{MFEA}, termed as Multi-Task Optimization~(MTO). The core design in solving MTO is the knowledge transfer scheme, which controls where, what and how the solving knowledge is shared across multiple optimization sub-tasks. Following this principle, Evolutionary Transfer Optimization (ETO)~\cite{ETO} has attracted considerable attention in recent years as a rapidly developing research topic. In particular, surrogate-assisted transfer optimization algorithms employ computationally cheaper surrogate models and augment them with knowledge from similar sources. Typical examples exist in the Bayesian optimization literature~\cite{MTGP_Bayesian,MTGP_Bayesian2,MTGP_Bayesian3}, as well as heterogeneous settings~\cite{Heter_TPO,HSVLMC}, with a comprehensive survey provided by Bai et al.~\cite{MTGP_Bayesian_survey}. Nevertheless, most of these works focus on single-objective optimization. To the best of our knowledge, limited works explore developing DDEA for multi-task-multi-objective optimization problems. In the limited literature, Luo et al.~\cite{GCS-MOE} introduced an intra-task transfer mechanism in decomposition-based evolutionary multi-objective optimization to share information among subregions of the Pareto front. Similar ideas also appear in~\cite{MOEA/D-MTGP}. Fan et al.~\cite{MOEA-D-EGO} combined transfer component analysis with MOEA/D-EGO to solve dynamic multi-objective optimization problems with expensive objectives. ExTrEMO~\cite{ExTrEMO} scalarizes a multi-objective problem into a set of single-objective subproblems and performs evolutionary transfer optimization by capturing positive or negative correlations between different source–target pairs. However, the surrogate models used in the above algorithms are all from the Gaussian process family; their inference time complexity grows with the amount of training data and the number of tasks, which limits their scalability.  Given the strong practical interest and urgent demand in MTMOO scenarios such as multi-objective VLSI physical design~\cite{MTMOO_EDA}, multi-objective alloy design~\cite{MTMOO_alloy}, MTMOO-oriented neural architecture search~\cite{MTMOO_NAS}, building performance design~\cite{MTMOO_building}, and multi-objective vehicle routing problems~\cite{MTMOO_routing}, more efficient and general DDEAs, particularly with novel surrogate modeling techniques, are highly desirable.

\subsection{LLM as Surrogate Model}
\label{subsec:motivation}
In this paper, we mainly consider MTMOO with offline setting. In such setting, surrogate modeling is still dominated by multi-task Gaussian processes (MTGPs). A common practice is to factorize along the objective dimension: either train a separate multi-task surrogate for each objective, or scalarize the multi-objective problem into a collection of single-objective subproblems and build an independent MTGP for each scalarized subproblem. ExTrEMO~\cite{ExTrEMO} follows the latter route. From a surrogate-modeling perspective, such designs have inherent limitations in offline, expensive settings. First, per-objective modeling learns only marginal mappings for individual objectives and cannot exploit shared latent structure across objectives within a single parameter space. Second, what truly matters in multi-objective optimization is the relative ordering and trade-off geometry in the objective space. When different objectives are predicted by independent surrogates, their errors can be misaligned in sign, scale, or bias, potentially altering dominance relations and distorting the Pareto front geometry. Per-objective training restricts knowledge transfer to a single objective channel and makes it difficult to capture cross-objective regularities—such as stable trade-off patterns shared within a task family—thereby fragmenting information along the objective dimension. These gaps naturally motivate a meta-surrogate that unifies multi-task and multi-objective modeling within a single model.

Against this backdrop, our prior work \emph{MetaSurrogate}~\cite{meta-surrogate} can be viewed as an initial exploration of using large language models (LLMs) as surrogates for data-driven evolutionary optimization. MetaSurrogate jointly encodes task metadata $m_t$ (e.g., function name, instance index, dimensionality), the decision vector $x$, and a \emph{single} objective value $y$ into a unified token sequence: metadata are templated as short natural-language prompts, while numerical decision variables and objective values are represented via \emph{scientific-notation encoding (SNE)}. For example, one input coordinate may be encoded as \verb|+ <10^0> 3 1 4 1 5|, and an objective value as \verb|- <10^3> 2 7 1 8 2 8|. During training, MetaSurrogate conditions on $(m_t, x)$ and minimizes a priority-weighted cross-entropy loss over the output token sequence with teacher forcing, demonstrating the feasibility of cross-task generalization in offline \emph{single-objective} settings. However, this route has three key limitations. \emph{First}, MetaSurrogate handles only single-objective modeling. \emph{Second}, pure teacher-forced, token-level cross-entropy introduces classic \emph{exposure bias}: at training time the model always predicts the next token under a \emph{gold} prefix, whereas at inference it must generate under its \emph{self-produced} prefix. For instance, suppose the ground-truth value is $1.23\times 10^3$ with prefix \verb|+ <10^3> 1 2 ...|; if at inference the model incorrectly emits \verb|<10^5>| for the exponent token, later mantissa tokens may look ``correct'' at the token level but the resulting number deviates by two orders of magnitude, and the error can compound along the generation process. \emph{Third}, the training loss is defined on \emph{per-token} cross-entropy, whereas the metric of interest is the error of the \emph{floating-point number} obtained by parsing the entire token sequence, yielding a typical loss–metric mismatch. Concretely, if $y=3.1415$ has SNE \verb|+ <10^0> 3 1 4 1 5|, the model may generate two candidates: $\hat y_1$ with \verb|+ <10^0> 3 1 4 1 6| and $\hat y_2$ with \verb|+ <10^0> 3 9 9 9 9|. From a token-level cross-entropy view, both differ from the reference at a few positions and thus incur similar CE losses; numerically, however, $(\hat y_1-y)^2 \ll (\hat y_2-y)^2$. Conversely, if we define a sequence-level numerical loss $L_{\mathrm{num}}=(\hat y-y)^2$, then $\hat y$ is obtained via a non-differentiable mapping (argmax/sampling followed by string parsing), which blocks gradient backpropagation. Such ``many tokens $\rightarrow$ one number with a single scalar loss'' makes it difficult to balance stability and metric consistency by relying solely on cross-entropy or by naively differentiating through the numeric error. In sequence learning (e.g., MT and image captioning), it is common to resort to reinforcement learning to optimize sequence-level metrics such as BLEU and CIDEr~\cite{Rennie_2017_CVPR,ziegler2020finetuninglanguagemodelshuman}. Analogously, in multi-task multi-objective surrogate modeling, remaining within a single-objective, token-level cross-entropy framework makes it hard to simultaneously mitigate exposure bias and loss–metric inconsistency.

Therefore, building on MetaSurrogate, we propose Q-MetaSur for expensive, offline MTMOO: under a unified textual representation, we introduce sequence-level \emph{numeric} rewards with offline Q-learning so that a single LLM jointly models multiple tasks and objectives while explicitly bridging the structural gap between \emph{token-level training objectives} and \emph{multi-objective numerical accuracy}.

\section{Proposed Method}
\label{sec:method}
In this section, we present Q-MetaSur, a LLM-based surrogate to assist DDEAs for solving MTMOO problems. Our main goal is to fine-tune a single LLM-based surrogate model that generalize across diverse sub-objective functions in MTMOO problems. In the rest of this section, we first provide a formal definition of such surrogate modeling problem, and then detail each novel designs that help our approach achieves desired generalization capability.

\subsection{Multi-Task Multi-Objective Surrogate Modeling}
\label{subsec:mm_surrogate}
We first define a MTMOO problem instance is a union of $T$ MOO sub-tasks: $\mathcal{T}:=\{\mathcal{T}_t\}_{t=1}^{T}$ where the task index is \(t\in\{1,\dots,T\}\). Then for $t$-th sub-task of this MTMOO instance, we define it as a four-element tuple: $\mathcal{T}_t:=(\mathcal{X}_t, F_t, \mathcal{M}_t,\mathcal{D}_t)$, of which each element describe a part of mathematical aspects of this sub-task and all contribute to a comprehensive and fine-grained task definition. Specifically, $\mathcal{X}_t$ gives the searching range of decision variables and $F_t$ gives the objective mapping that maps a decision variables vector $x$ to a multi-objective space:
\begin{equation}
    F_t(x)\triangleq\big(f_t^{(1)}(x),\ldots,f_t^{(k_t)}(x)\big)\in\mathbb{R}^{k_t},
\quad f_t^{(j)}:\mathcal{X}_t\to\mathbb{R}.
\end{equation}
where $k_t$ denotes the number of objective functions in this MOO sub-task. The other two elements bring concepts of LLM-based surrogate modeling and offline surrogate setting into the context. In specific, the metadata $\mathcal{M}_t$ is a json-like textual description for profiling the features of the sub-task, e.g., the task name, dimension and etc. The offline evaluation data $\mathcal{D}_t$, on the other hand, provides fine-grained regression feature of the sub-task through a collection of $N_t$ historical evaluation data:
\begin{equation}
    \mathcal{D}_t=\{(t,m_t,x_i^{(t)},\mathbf{y}_i^{(t)})\}_{i=1}^{N_t},
\quad \mathbf{y}_i^{(t)}=F_t(x_i^{(t)})\in\mathbb{R}^{k_t}.
\end{equation}
where we additionally include the task index $t$ and the metadata $m_t$ within the offline dataset to facilitate our subsequent language modeling-based surrogate learning. Given all elements being precisely defined, suppose we have a surrogate model at hand~(with learnable parameters $\theta$), then we can summarize the learning objective of MTMOO surrogate modeling problem as belows:
\begin{equation}
\begin{split}
J(\theta) = \mathbb{E}_{x\in \mathcal{D}} \left[\frac{1}{2}||\hat F_\theta(x) - F_{t(x)}(x)||_2^2\right].
\end{split}
\end{equation}
where $\mathcal{D}=\cup_{t=1}^{T}\mathcal{D}_t$ is the union of offline datasets of all sub-tasks, $t(x)$ is an indexing function that tells a data item $x$ belongs to which sub-task. In this paper, we aims to minimize $J(\theta)$ hence we could use a universal and generalizable surrogate $\theta$ to provide accurate surrogate modeling for diverse objective functions within a MTMOO problem instance. Specifically, our Q-MetaSur is parameterized as a conditional language model:  
\begin{equation}
p_{\theta}(\mathbf{y}\mid m_t,x)\ \equiv\ p_{\theta}\big(\tau(\mathbf{y})\mid \tau(m_t,x)\big),
\end{equation}
where \(\tau(\cdot)\) maps metadata, decision vectors, and objective vectors into unified token sequences (the LLM's tokenizer); \(\tau(\mathbf{y})\) is the token sequence of the multi-objective output. Training first aligns the sequence-level conditional distribution via maximum likelihood and then, in the offline setting, further optimizes via offline Q-learning under a reward sensitive to numerical error, thereby improving continuous regression accuracy. To handle variable $k_t$, we use ``objective delimiters'' and an ``end-of-sequence'' token to bound outputs and generate only up to the $k_t$-th objective; this aligns with the textual representation and naturally adapts to varying $k_t$. An important note is that the learning objective we summarized above provides a strong motivation to prompt us adopting large language model as effective surrogate modeling tool. The reasons are in two-folds: 1) Considering that diverse objective functions in diverse MTMOO problems, we need a model with high capacity to learn across various landscape knowledge; 2) LLM provides unified textual sequence representation (see \autoref{subsec:mm_text_repr}) to align heterogeneous \(\{n_t\}\) and \(\{k_t\}\) at the sequence level. Next we delve into the technical details.

\subsection{Textual Representation}
\label{subsec:mm_text_repr}
In Q-MetaSur, we provide a unified textual representation for \((m_t,x,\mathbf{y})\) so that heterogeneous decision dimensions \(n_t\) and numbers of objectives \(k_t\) are handled consistently by a single model at the sequence level, enabling knowledge sharing across tasks and objectives within the surrogate. 

\subsubsection{Scientific Notation Encoding}
For any scalar value \(z\in\mathbb{R}\), let \(\mathrm{sign}(z)\in\{+,-,0\}\). When \(|z|>0\), set
\[
\kappa=\big\lfloor \log_{10}|z|\big\rfloor,\quad |z|=m\cdot 10^\kappa,\ \ m\in[1,10).
\]
With \(n_{digit}\) significant digits, SNE maps \(z\) to the token sequence
\begin{equation}
  \phi(z)=[\pm]\;\langle 10^{\kappa}\rangle\; d_1 d_2 \ldots d_{n_{digit}},
\end{equation}
where $[\pm]$ is the sign, $\kappa$ is the exponent for the most significant digit $d_1$, \(\langle 10^{\kappa}\rangle\) is the exponent token, and $d_1\ldots d_{n_{digit}}$ are the \(n_{digit}\)-digit mantissa with \(d_i\in\{0,\dots,9\}\). 
\subsubsection{Templated Metadata}
\begin{figure}[htbp]
      \centering
    \begin{promptmini}[colframe=green!60!black, colback=green!5]{metadata}
    \label{prompt_Small}
    \footnotesize
    You are a multi-objective multi-task surrogate model, predict fitness given m and pop;
    function name is \{$f_{\mathrm{name}}$\},\;
    function ID is \{$f_{\mathrm{ID}}$\},\;
    key feature 1 is \{$k_1$\}\,|\,key feature 2 is \{$k_2$\}\,|\,\dots\,|\,
    the dimensionality is dim=\{$D$\}.
    \end{promptmini}
  \caption{Structured template for the metadata.}
  \label{fig:metadata}
\end{figure}
For cross-task generalization, the textual metadata $m_t$ must convey \emph{what} is being optimized (function identifier) and \emph{how} it is parameterized (dimension and other instance-specific features). We design the structured template in \autoref{fig:metadata} with vertical bars as optional-field separators. The tuple $(f_{\mathrm{name}},f_{\mathrm{ID}},k_1,\dots,k_L,D)$ is rendered as a single self-contained sentence whose fields are separated by ``|'' to minimize grammatical ambiguity. The function ID (e.g., $F\!1$--$F\!24$ in the CEC2019 suite) serves as an unambiguous identifier when function names may collide across libraries, while free-form ``key features'' (e.g., ``CEC2019'', ``instance=0'') allow inserting arbitrary domain descriptors without changing syntax. The ellipsis in the prompt indicates that additional tokens can be transparently inserted, making the scheme extensible to richer contextual prompts in practice (e.g., noise level, budget category, simulator version). Furthermore, any string that uniquely identifies the problem is acceptable, not necessarily a function name; for example, a data distribution or a fitness-landscape fingerprint, or even a problem index.

\subsubsection{Serialization of the Decision Vector \(x\) and Multi-Objective \(\mathbf{y}\)}
The decision vector of length \(n_t\) is SNE-encoded elementwise and enclosed in square brackets, separated by commas:
\[
\Phi(x)=[\,\phi(x^{(1)}),\ldots,\phi(x^{(n_t)})\,].
\]
The multi-objective output \(\mathbf{y}=(y^{(1)},\ldots,y^{(k_t)})\) is likewise SNE-encoded elementwise:
\begin{equation}
\Phi(\mathbf{y})=[\,\phi(y^{(1)}),\ldots,\phi(y^{(k_t)})\,].
\end{equation}

\subsubsection{Example}
The representation is consistent for training and inference: the encoder takes the textual sequence of \((m_t,x)\), and the decoder generates the textual sequence of \(\mathbf{y}\). Because each objective appears as an independent SNE segment, the model can learn the joint distribution of objectives with different magnitudes at the token level.

\begin{figure}[htbp]
    \centering
    \includegraphics[width=0.8\linewidth]{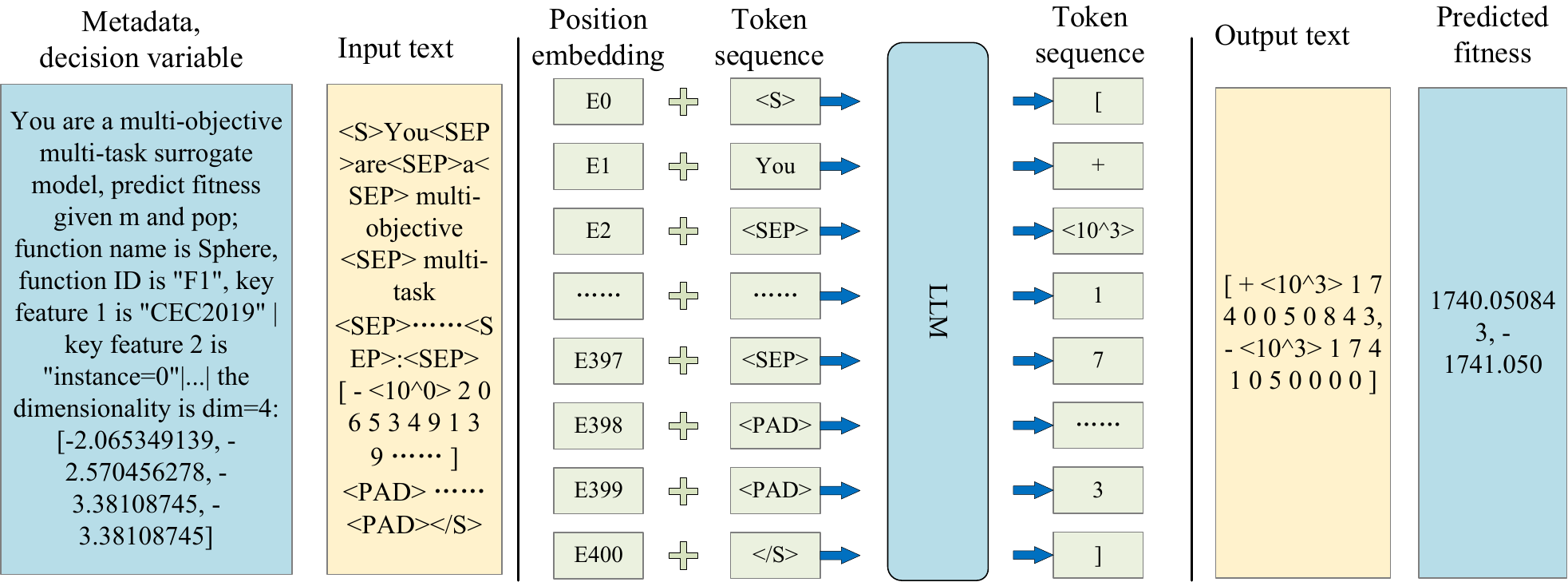}
    \caption{LLM-based prediction of decision variables and objective values.}
    \label{fig:dy}
\end{figure}

\autoref{fig:dy} illustrates the conversion pipeline. For the four-dimensional Sphere function (F1), the original description
\lstinline|CEC2019, instance=0, Sphere, dimension=4|
is converted into a prompt whose compact, compositional form allows the encoder to localize task semantics with negligible prompt overhead, avoiding the cumbersome feature mapping and joint-kernel engineering typical of multi-task GPs. At inference time, the metadata string and the decision vector $x$ in SNE are concatenated and fed to the encoder--decoder LLM, which decodes the multi-objective fitness $\mathbf{y}$ in the same textual format. This unified representation of $(m_t,x,\mathbf{y})$ enables a single conditional language model $p_\theta(\mathbf{y}\mid m_t,x)$ to learn joint embeddings over heterogeneous tasks without retraining a task-specific model and supports knowledge transfer across tasks and objectives. The revised metadata thus provides an extensible, machine-readable interface connecting heterogeneous optimization tasks with the LLM, while the consistent token-level granularity across $m_t$, $x$, and $\mathbf{y}$ preserves pretrained token semantics that underwrite effective transfer in the surrogate.

\subsection{Supervised Fine-Tuning}
We now elaborate how we train LLM-based surrogate model in the proposed representation system, with offline setting and generalization requirement. For a training sample $(t,m_t,x,\mathbf{y})$ with $\mathbf{y}=F_t(x)$, let the tokenizer be $\tau(\cdot)$ and define $Z:=\tau(m_t,x)=(z_1,\dots,z_{n_{\text{src}}})$ and $O:=\tau(\mathbf{y})=(o_1,\dots,o_{n_{\text{tgt}}})$, where $n_{\text{src}}=|Z|$ and $n_{\text{tgt}}=|O|$. Let the encoder and decoder share the embedding matrix $\mathbf{E}$~(co-trained with LLM's parameters) and embed tokens as LLM's input:
\[
X \;=\; \mathrm{PE}(1{:}n_{\text{src}})\;+\;
[\mathbf{E}(z_1);\dots;\mathbf{E}(z_{n_{\text{src}}})]
\ \in\ \mathbb{R}^{n_{\text{src}}\times d_m},
\]
where $\mathrm{PE}(\cdot)$ is the positional encoding aligned with $X$. Denote the encoder output by $\mathrm{Encoder}(X)$, and use teacher forcing in the decoder to generate $O$ autoregressively. The negative log-likelihood (token-level cross-entropy) is
\begin{equation}\label{eq:nll}
\mathcal{L}_{\mathrm{NLL}}
\;=\;
-\sum_{i=1}^{n_{\text{tgt}}}
\log P_{\theta}\bigl(o_i \mid o_{<i},\,\mathrm{Encoder}(X)\bigr),
\end{equation}
with $o_{<i}:=(o_1,\dots,o_{i-1})$. However, in our preliminary experiments, we found that let the NLL loss focus evenly on each output token may impact the training effectiveness. This is due to that, given the SNE structure of $\tau(\mathbf{y})$, each objective forms a segment in the label sequence, containing the sign, the exponent token, and the mantissa digits. To explicitly reflect positional importance for numerical semantics, we employ a \emph{priority-weighted cross-entropy} (PWCE) within each segment. To achieve this, we first partition the label output token sequence into $k_t$ segments (each objective in the MOO sub-task is a segment). Then, we use a decayed weight to force the NLL loss focus more on the sign, exponent and previous mantissa part of the output token sequence, since these parts hold more impact on the accuracy of the overall objective prediction, the attained PWCE loss can then be formulated as:
\begin{equation}
\begin{split}
\mathcal{L}_{\mathrm{PWCE}}
= -\sum_{j=1}^{k_t}\sum_{\ell=1}^{L_j}\omega_{j,\ell}\,
   \log P_{\theta}\!(
      o_{\iota_{j,\ell}} \,|\, o_{<\iota_{j,\ell}},
 \ \, \mathrm{Encoder}(X)
   ).
\end{split}
\end{equation}
\begin{equation}
\omega_{j,\ell}=
\begin{cases}
20, & \ell\in\{1,2,3\},\\[0.2em]
\max\!\bigl(1,\,10-(\ell-4)\bigr), & \ell\ge 4,
\end{cases}
\end{equation}
where $k_t$ is the number of objectives in the $t$-th MOO sub-task and $L_j$ is the number of tokens used for tokenizing each objective value. As mentioned above, $\omega_{j,\ell}$ denotes the weight we use to trade off the attention of our loss function.
We use this loss to fine-tune a pretrained LLM under the offline data collected from MTMOO problem instances, in a regular supervised learning routine.

\subsection{Offline Q-Learning Fine-Tuning}
\begin{figure}[htbp]
    \centering
    \includegraphics[width=0.8\linewidth]{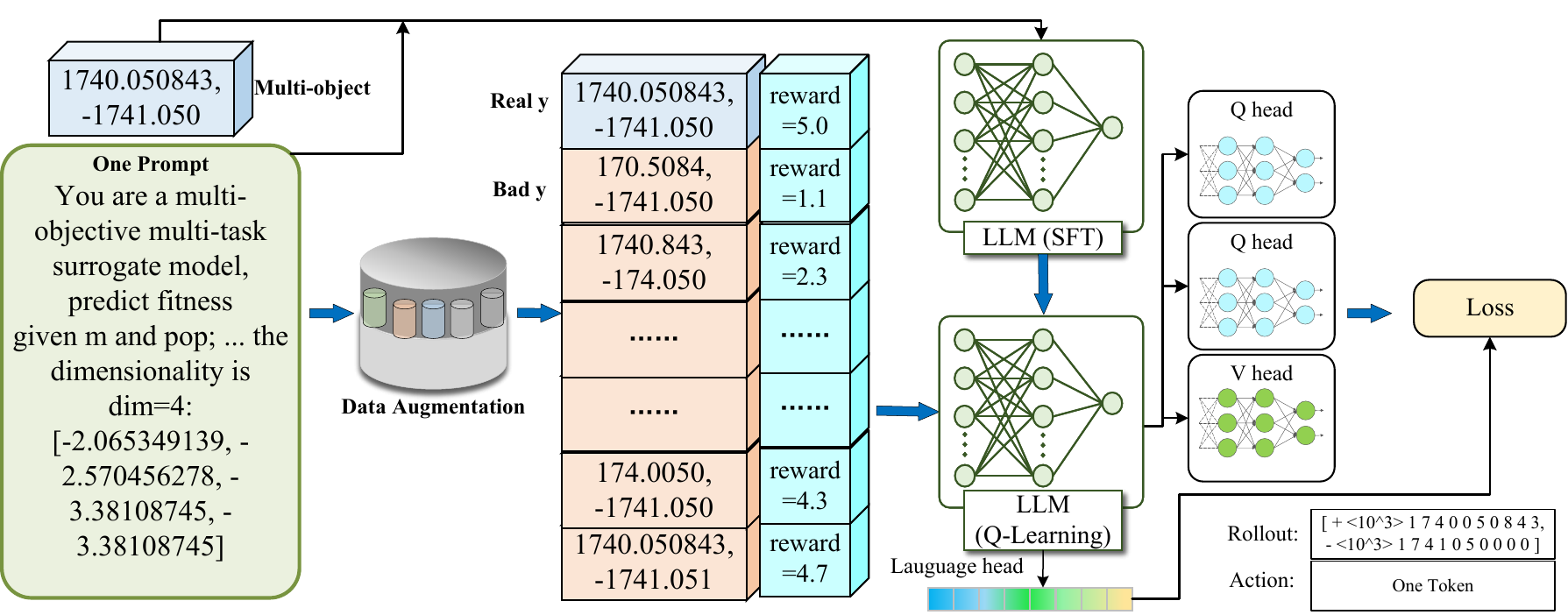}
    \caption{Two-stage fine-tuning architecture (illustrated for the bi-objective case). In the first stage, supervised fine-tuning learns the token sequences corresponding to the correct pair of objective values for each training sample. In the second stage, offline Q-learning fine-tuning first constructs and augments a large number of “incorrect” objective sequences around the ground-truth labels and computes numerical rewards for them in an offline manner, thereby alleviating the scarcity and narrow support of the offline data and substantially increasing the diversity of both data and rewards; it then performs value learning on the augmented dataset. In contrast, supervised fine-tuning only fits the correct answers, whereas offline Q-learning further enables the model to characterize the degree of error associated with different incorrect predictions.
}
    \label{fig:SFTQ}
\end{figure}
While supervised fine-tuning~(SFT) can let the LLM quickly fit the offline evaluation data, it may also introduce overfitting risk, which subsequently heavily impacts the generalization capability of the LLM-based surrogate onto unseen problem instances. A common practice in existing LLM fine-tuning literature is to introduce an additional reinforcement learning~(RL) stage~\cite{Rennie_2017_CVPR,ziegler2020finetuninglanguagemodelshuman} to guide the model with exploratory trajectories that ensure more comprehensive learning. In our work, we follow this paradigm and propose a tailored offline RL workflow based on ILQL to endow the LLM-based surrogate with more generalizable prediction ability. Intuitively, SFT is responsible for learning the basic “language + numeric expression’’ around the offline samples, whereas the subsequent offline ILQL stage, driven by sequence-level rewards and conservative $Q/V$ estimates, moves the model from merely imitating labels to actively exploring the neighbourhood of the ground truth and biasing generation towards numerically more accurate and stable outputs. As summarized in Fig.~\ref{fig:SFTQ}.

Following the token-level POMDP formulation used in ILQL~\cite{ILQL}, under the unified textual representation, a single decoding pass is treated as a finite-horizon partially observable Markov decision process: given the input sequence \(\tau(m_t,x)\), the decoder generates the target sequence \(\tau(\hat{\mathbf{y}})=(z_1,\ldots,z_L)\) step by step. We model this as an episodic POMDP
\(\mathcal{M}=(\mathcal{S},\mathcal{O},\mathcal{A},P,\Omega,r,\rho_0,\gamma,H)\).
Here, \(\mathcal{S}\) denotes latent environment states that collect the MTMOO instance \((t,m_t,x,\mathbf{y})\) and the token index; the agent only observes token histories \(h_i=(\tau(m_t,x),z_{<i})\) from the observation space \(\mathcal{O}\).
The action space
\(\mathcal{A}\subseteq\mathcal{V}\) is the tokenizer's \emph{numeric-support} subset (signs \(\{+,-\}\), exponent tokens \(\langle 10^{k}\rangle\), digits \(0\)--\(9\), and required separators/terminators). Within an episode, the transition kernel \(P\) deterministically appends the chosen token to the history, and the observation kernel \(\Omega\) reveals the updated prefix; generation terminates at EOS (or \texttt{]}), or when reaching the length limit \(H_{\max}\), yielding episode length \(L=|\tau(\hat{\mathbf{y}})|\le H_{\max}\). The reward is the sequence-level return \(R(\hat{\mathbf{y}};\mathbf{y})\), credited to the final token.

\subsubsection{Data Augmentation}
\label{data_augmentation}
Offline datasets are typically expensive, scarce, sparse, and narrowly distributed. Training solely on ground-truth samples leads to limited reward diversity and poor generalization.To address this, we design a set of complementary perturbation mechanisms applied to $\mathbf{y}$.
In our work, augmentation is conducted by adding different level of noises into the ground truth offline data. Specifically, we do not alter the input data to the LLM-base surrogate model, instead, we alter the predicted objective value with noise to produce moderate ``bad'' data. These false data can be regarded as additional error information~(feedback) that stabilize the RL fine-tuning. To provide a comprehensive data augmentation, we propose three noise fusion strategies to add low, medium, and high-level noises to the original offline data. The low-level noise considers adding small scale gaussian noise into the label output objective value directly: $y^{(j)}=y^{(j)}+\xi^{(j)}$. The medium-level noise considers perturbing the objective values in  logarithm magnitude: $y^{(j)}=\mathrm{sign}(y^{(j)})\cdot 10^{\,\log_{10}\!\lvert y^{(j)}\rvert+\epsilon^{(j)}}$. The high-level noise considers shifting the objective values with a significant portion of its range: $y^{(j)}\sim \mathcal{U}\!(y_{\min}^{(j)}-\delta \Delta_j,\ y_{\max}^{(j)}+\delta \Delta_j)$, where $\Delta_j$ is the range of the objective values. By using these noise-adding strategy, we construct a augmented dataset with noisy evlauation data, which is termed as $\hat{D}$. Our Rl-based fine-tuning operates on the union of the tral evaluation dataset $D$ and such noisy dataset $\hat{D}$. We leave concrete settings at experiment sections.

\subsubsection{Reward Definition}
Another important design in our work is how to credit proper RL feedback signal to LLM-based surrogate model to facilitate effective RL fine-tuning. To this end, we specifically design a reward function that is capable of precisely reflecting the surrogate model's accuracy across diverse sub-tasks and diverse objective functions in MTMOO problems.  We adopt a bounded, small-error-sensitive exponential of RMSE as the main reward, plus small SNE-based exponent/sign bonuses to reflect the learning path of ``align sign and exponent first, then refine mantissae''. Let $\Delta_j=y_{\max}^{(j)}-y_{\min}^{(j)}$ be the range per objective in a task and define the per-task, per-objective normalized RMSE:
\[
\mathrm{nRMSE}=\sqrt{\frac1{k_t}\sum_{j=1}^{k_t}
\Bigl(\frac{\hat y^{(j)}-y^{(j)}}{\Delta_j}\Bigr)^2}.
\]
The reward is then computed as a weighted sum of nRMSE and two additional bonus signals:
\begin{equation}
\label{eq:reward}
\begin{aligned}
R(\hat{\mathbf{y}};\mathbf{y})
&= \exp\!\Big(-\frac{\mathrm{nRMSE}(\hat{\mathbf{y}},\mathbf{y})}{s}\Big) \\
&\quad + \kappa_{\exp}\,\mathrm{Acc}_{\exp}(\tau(\hat{\mathbf{y}}),\tau(\mathbf{y})) \\
&\quad + \kappa_{\mathrm{sgn}}\,\mathrm{Acc}_{\mathrm{sgn}}(\tau(\hat{\mathbf{y}}),\tau(\mathbf{y})),
\end{aligned}
\end{equation}
where $s$ is a temperature (default $0.03$) that scales the dimensionless error in $R_{\text{main}}=\exp(-\mathrm{nRMSE}/s)$, thereby controlling how fast the reward decays with error: smaller $s$ imposes stronger penalization and higher sensitivity near small errors, whereas larger $s$ yields a flatter, more tolerant curve. 
\(\mathrm{Acc}_{\mathrm{sgn}}\) are per-objective exponent/sign matching rates (parsed from SNE segments). The small weights \(\kappa_{\exp},\kappa_{\mathrm{sgn}}\) default to \(0.2\) and \(0.05\). This reward offers large gradients near small errors, an upper bound, and structural consistency with SNE; in practice, we clip \(R\) to $[0,\ 5]$ to avoid domination by extreme candidates. Such a fine-grained reward design ensure stable learning.

\subsubsection{Training}
In the offline RL setting of this paper,  training language model to predict surrogate objective value in a sequence to sequence fashion poses certain challenges for training the LLM model. Unlike our proposed first-stage supervised fine-tuning where each output token has corresponding supervision signal, in this offline RL fine-tuning stage, we have to address two key issues: 1) as we mentioned previously, use sequence-to-sequence model to auto-regressively predict tokens of the target objective values is a POMDP, which require careful designs to ensure learning stability; 2) training RL over offline dataset often leads to distribution shift issue, which further results in over-estimation on unseen data. To address these two key issues, we in this paper introduce a synergy of Implicit Language Q-Learning~(ILQL)~\cite{ILQL} and Conservative Q-Leaning~(CQL)~\cite{CQL} to ensure stable fine-tuning in this stage.

For the first issue, following ILQL, we learns both Q and V values from the offline dataset. Specifically, after the supervised fine-tuning stage, we add two Q heads and a V heads to the LLM neural network architecture. For the presentation simplicity, we use $Q_\theta$ to denote the behavior Q head, $Q_{\hat{\theta}}$ to denote the target Q head, $V_\theta$ to denote the V head. Then given the offline evaluation dataset $\mathcal{D}$, we learns these value heads by bootstrapping each other:
\begin{equation}
    \mathcal{L}_{Q,V} = \mathbb{E}_{\tau \in \mathcal{D}\cup \hat{\mathcal{D}}}\left[\sum_{i=1}^{L}(R(h_i, a_i) + \gamma * V_\theta(h_{i+1})-Q_\theta(h_i,a_i))^2  + ET(Q_{\hat{\theta}}(h_i,a_i)-V_\theta(h_i))\right].
\end{equation}
where $\tau$ is the training data in both the real evaluation dataset and our augmented dataset, $h_i$ denotes the hidden feature vector the LLM outputs for i-th token, $ET(\cdot)$ denotes the expectile loss proposed in IQL~\cite{IQL} to learn implicit maximization of V heads through bootstrapping. During training, the LLM's parameters are co-trained with these three value heads. This learning objective stabilizes the RL fine-tuning by token-level Bellman update.

For the second issue, a common observation in existing offline RL literature~\cite{Offline_RL_overview} is that, if we use normal Bellman update to train the RL agent on the offline dataset, it unavoidably results in over-estimation of value prediction on out-of-distribution~(OOD) data. This would significantly impact the performance of our LLM-based surrogate model, since a single token shift may cause the predicted objective varies by orders of magnitude. Hence, we leverage conservative Q value regularization trick in CQL~\cite{CQL} to mitigate OOD shift by reshaping the predicted token distribution to fit the training offline data. The added regularization term can be formulated as:
\begin{equation}
\label{eq:cql_ce}
\mathcal{L}_{\mathrm{CQL}}
=\mathbb{E}_{(s,a)\in \hat{\mathcal{D}}\cup \mathcal{D}}\Big[\ \mathrm{CE}\big(\mathrm{softmax}(Q_\theta(s,\cdot)),\ a\big)\ \Big].
\end{equation}
where $CE(\cdot)$ operates in a cross-entropy loss fashion and pushes the Q value head to maximize the q value at the sampled action in offline data. This would significantly decrease the risk of over-estimation on other action slots that have not been sampled by the behavior policy. We have to note that, despite the RL loss we have introduced above, we additionally include a memorization forcing loss term to restrict the learning update obtained from RL training, which is the normal negative-log-likelihood loss $\mathcal{L}_{NLL}$ in language model training. It is the simple cross-entropy in Eq.~\ref{eq:nll}. In our preliminary experiment, we found adding such revision term would help LLM learns better prediction policy while keeps the learned surrogate modeling knowledge in the first supervised fine-tuning stage. The final learning objective is formulated accordingly as below:
\begin{equation}
\label{eq:total}
\mathcal{L}
=\mathcal{L}_{Q,V}+\lambda_{\mathrm{CQL}}\mathcal{L}_{\mathrm{CQL}}+\mathcal{L}_{\mathrm{NLL}}
\end{equation}

\begin{algorithm}[t]\algsetup{linenosize=\tiny}\small
\caption{MO\_MaTDE with the Q-MetaSur.}
\label{algo:MO_MaTDE_LLM}
\begin{algorithmic}[1]
\REQUIRE $im$ \COMMENT{migration prob.} ; \textit{aUp} \COMMENT{archive update prob.} ; \textit{shk} \COMMENT{shrink factor}
\ENSURE Approximated Pareto sets/fronts of all tasks
\STATE LLM $\gets$ fine-tuned LLM (Supervised Fine-tuning + Offline Q-learning Fine-tuning)
\STATE Initialize $\{\textit{pop}[t]\}_{t=1}^T$ (size $N$), reward $\textit{rew}\in\mathbb{R}^{T\times T}\leftarrow\mathbf{1}$, possibility $\textit{pos}\leftarrow\mathbf{0}$, archives $\{\textit{arc}[t]\}_{t=1}^T$
\FOR{$t=1$ \TO $T$} \STATE $\textit{pop}[t]\gets\textsf{NSGA2\_Sort\_Trunc}(\textit{pop}[t],N)$ \ENDFOR
\WHILE{termination not met}
  \FOR{$t=1$ \TO $T$}
    \IF{$\mathrm{rand()} > im$} 
      \STATE Randomize $F,CR$ for each individual in $\textit{pop}[t]$
      \STATE $\textit{offs}\gets\textsf{DE\_Generate}(\textit{pop}[t])$
      \FOR{each $u\in\textit{offs}$} \STATE $u.\textsf{Obj}\gets\hat F_{\mathrm{meta}}(m_t,u.\textsf{Dec})$ \COMMENT{LLM inference (SNE-encoded)} \ENDFOR
      \STATE $\textit{pop}[t]\gets\textsf{NSGA2\_Select}(\textit{pop}[t]\cup\textit{offs},N)$
    \ELSE 
      \STATE $\textit{tfTsk}\gets\textsf{AdaptiveChoose}(t,\textit{arc},\textit{rew},\textit{pos})$
      \STATE $\textit{offs}\gets\emptyset$
      \FOR{each $x\in\textit{pop}[t]$}
        \STATE $\mathrm{donor}\sim\textit{pop}[\textit{tfTsk}]$; $u\gets\textsf{DE\_Crossover}(x,\mathrm{donor})$; add $u$ to \textit{offs}
      \ENDFOR
      \FOR{each $u\in\textit{offs}$} \STATE $u.\textsf{Obj}\gets\hat F_{\mathrm{meta}}(m_t,u.\textsf{Dec})$ \COMMENT{LLM inference (SNE-encoded)} \ENDFOR
      \STATE $\textit{pop}[t]\gets\textsf{NSGA2\_Select}(\textit{pop}[t]\cup\textit{offs},N)$
      \STATE \textbf{if} best-of-$\textit{pop}[t]$ improved \textbf{then} $\textit{rew}[t,\textit{tfTsk}]\gets\textit{rew}[t,\textit{tfTsk}]/\textit{shk}$ \textbf{else} $\textit{rew}[t,\textit{tfTsk}]\gets\textit{rew}[t,\textit{tfTsk}]\times\textit{shk}$
    \ENDIF
    \FOR{each $x\in\textit{pop}[t]$} \IF{$\mathrm{rand()}<\textit{aUp}$} \STATE insert $x$ into $\textit{arc}[t]$ (random replace if full) \ENDIF \ENDFOR
  \ENDFOR
\ENDWHILE
\STATE \textbf{Output:} non-dominated sets/fronts of all tasks
\end{algorithmic}
\end{algorithm}

\subsection{Inference with the Fine-Tuned LLM}
\begin{figure}[htbp]
    \centering
    \includegraphics[width=0.8\linewidth]{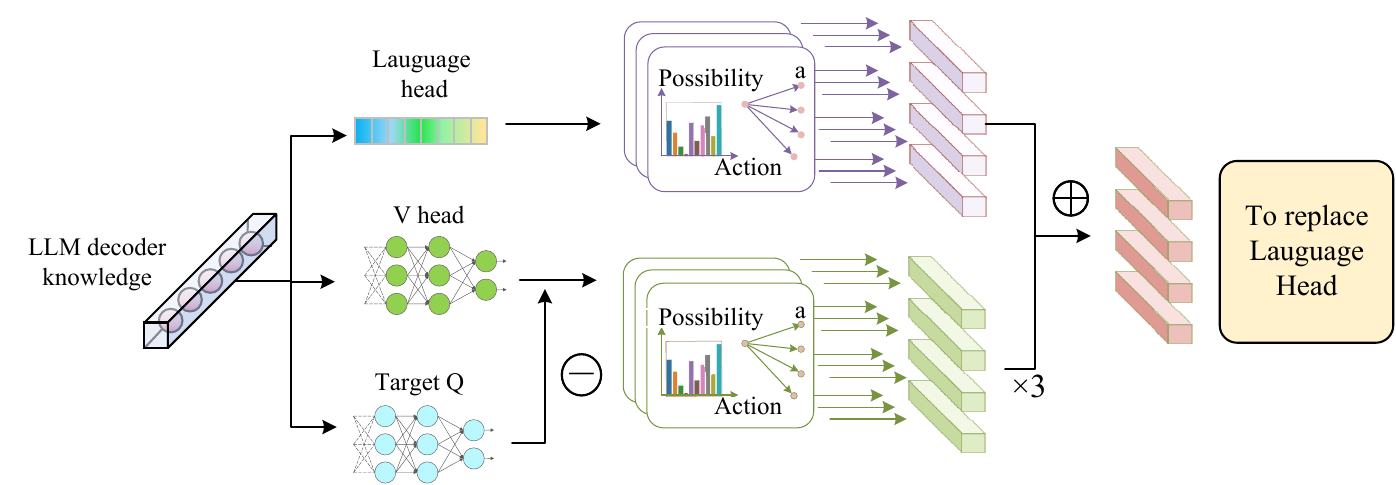}
    \caption{Advantage-guided inference mechanism. For each input sample, the model is forward-propagated to the language modeling head of the last decoder layer, where the advantage of each candidate token is computed using \((\text{target } Q - V)\), scaled (multiplied by 3), and added to the corresponding logits. Tokens with positive advantage thus obtain higher selection probabilities, whereas those with negative advantage are suppressed. The subsequent decoding process is identical to that of a standard language model and can employ greedy decoding or top-$k$ sampling to generate the target sequence.
}
    \label{fig:infer}
\end{figure}

\begin{figure}[htbp]
    \centering
    \includegraphics[width=0.3\linewidth]{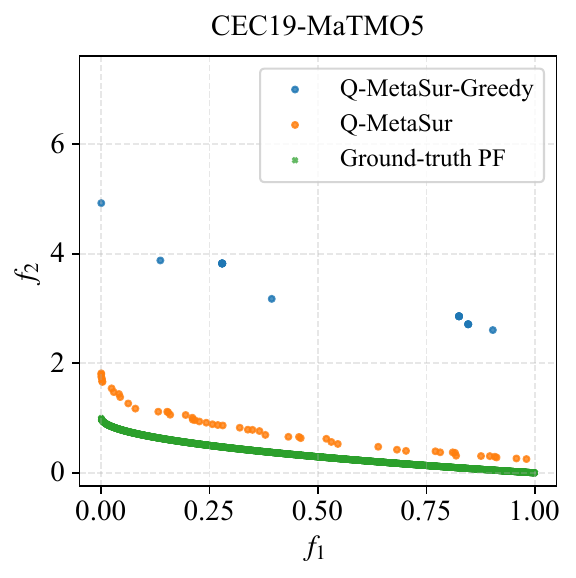}
    \caption{Comparison of the ground-truth Pareto front, Q-MetaSur, and Q-MetaSur-Greedy (without RL-based advantage-guided inference). Under identical backbone LLM weights, the proposed advantage-guided inference in Q-MetaSur yields a Pareto front that more closely matches the ground-truth front than greedy decoding.}
    \label{fig:infer_greedy}
\end{figure}

After offline RL fine-tuning, the model is deployed for inference. We keep the backbone LLM fixed and, at each decoding step, apply an advantage-based correction to the log probabilities to improve stability and reliability.
Concretely, for each token, we obtain an adjusted likelihood by adding its estimated advantage to the original logits of the LLM and scaling this term by a multiplicative factor (set to 3 by default).
This procedure amplifies the probabilities of tokens with positive advantage and suppresses those with negative advantage. Fig.~\ref{fig:infer} depicts the proposed advantage-guided inference procedure for the LLM. As shown in the ~\ref{fig:infer_greedy}, with the large-model weights kept exactly the same, our inference strategy achieves a significantly better Pareto front than greedy sampling.

\subsection{Integration of Q-MetaSur with MTMOO Algorithms}
Our meta-surrogate integrates seamlessly with any MTMOO algorithm to assist expensive fitness evaluations. Taking MaTDE~\cite{MaTDE} as an example, its pseudocode is in \autoref{algo:MO_MaTDE_LLM}. The meta-surrogate predicts pseudo-fitness (Lines 12 and 22). The method suits scenarios with expensive evaluations or offline settings where only sampled data points are available without explicit evaluators. Notably, offline DDEAs can be integrated into online DDEAs as well: they can provide a solid initialization for online algorithms or serve as a pre-processing step for new data-driven optimization tasks, improving overall performance and efficiency.

\section{Experiments}
\label{sec:experiments}
In this section, we investigate three research questions. (RQ1) \emph{Surrogate feasibility}: after training on tokenized decision variables, fitness values, and task metadata, can a large language model serve as a meta-surrogate for multi-objective, cross-task fitness prediction? (RQ2) \emph{Emergent capability}: to what extent can the meta-surrogate generalize to tasks beyond its training distribution, especially those with unseen input dimensions and unseen task parameterizations? (RQ3) \emph{Performance evaluation}: when the LLM surrogate is integrated into the MTMOO framework, how effectively does it enhance evolutionary search?

\subsection{Surrogate Evaluation (RQ1)}

\subsubsection{Problem Setup}
We adopt T5 as the backbone language model. The multi-objective, multi-task benchmark was proposed in the IEEE CEC 2019 competition on evolutionary multitask optimization. It comprises six multi-objective, 50-task suites constructed by applying different rotation matrices to ZDT or DTLZ problems. The test suites contain six instances, each instance comprising 50 tasks; implementations are available in the MTO-Platform\footnote{\url{https://github.com/intLyc/MTO-Platform}}. 

For classical surrogates, one typically trains a separate model \emph{per objective and per problem}, whereas a multi-task Gaussian process (GP) can jointly train across tasks for each objective. In contrast, our LLM trains a \emph{single} conditional model that covers all tasks and dimensions via prefix tuning. The train/test split is \textit{5:3}, and each task provides 200 training samples.

Across all experiments, we report the \emph{scaled mean absolute error} (sMAE; min–max normalized within each task) and the coefficient of determination ($R^2$)\footnote{We compute $R^2$ using \texttt{scikit-learn}; its theoretical range is $(-\infty,\,1]$.} as the primary metrics. To remove scale differences between tasks, sMAE normalizes the absolute error within each task. Let ${I}_k^{\text{test}}$ denote the index set of test samples for task $k$. Let $y_i^{k}$ be the ground-truth fitness and $\hat y_i^{k}=\hat f_{\mathrm{meta}}(x_i^{k},m_k)$ the predicted fitness. Define the task-wise value range
\[
R_k=\max_{j\in I_k^{\mathrm{test}}} y_j^{k}
    -\min_{j\in I_k^{\mathrm{test}}} y_j^{k}.
\]
Then the per-task sMAE (and its overall mean) is
\begin{equation}
    \mathrm{sMAE}\;=\;\frac{1}{|{I}_k^{\text{test}}|}
    \sum_{i\in{I}_k^{\text{test}}}
    \frac{\lvert\,y_i^{k}-\hat y_i^{k}\,\rvert}{R_k}.
\end{equation}

\subsubsection{Training Setup}
For the LLM, we use publicly available pretrained checkpoints from Hugging Face\footnote{\url{https://huggingface.co/}}. The input sequence (metadata concatenated with numeric tokens) has \(20\times D\) tokens, and the target sequence length is 50 tokens. The LLM is trained for 65 epochs with supervised fine-tuning (SFT) and for 5 epochs with offline Q-learning. Experiments run on a machine with an Intel Xeon E5-2680 v4 CPU, an NVIDIA GeForce RTX 4090 (24~GB) GPU, and 128~GB RAM. The code is implemented in Python with PyTorch~1.8.2. Table~\ref{tab:training_config} summarizes key hyperparameters, optimizer settings, and other training configurations.

Our value function is implemented by attaching three MLP heads to the last hidden layer of the T5 model: two independently initialized and separately trained Q-heads and one V-head. Each head consists of two layers, with a hidden dimension that is twice the embedding dimension. The target Q-value is parameterized as the minimum of the predictions from the two Polyak-averaged target Q-heads, i.e., $\hat{Q} = \text{min}(Q_1, Q_2)$.

During the offline Q-learning stage, we activate the perturb-and-score augmentation scheme. For each ground-truth sample $(t,m_t,x,\mathbf{y})$, we additionally generate 150 perturbed label vectors, consisting of 50 low-noise, 50 medium-noise, and 50 high-noise variants. To keep the three noise levels consistent with the method description, low-level noise is implemented as small zero-mean Gaussian perturbations whose standard deviation is set to approximately $1\%$ of the empirical range of each objective within the corresponding task, so that only the mantissa and low-order digits are slightly shaken without changing the overall order of magnitude. Medium-level noise operates in the logarithmic (base-10) magnitude domain with zero-mean Gaussian perturbations of standard deviation $0.15$, which roughly corresponds to randomly scaling each objective value within about $0.7$–$1.4$ times its original magnitude and thus enables moderate exploration around the ground truth in terms of scale. High-level noise is realized by uniformly sampling within an extended interval obtained by enlarging the empirical minimum and maximum of each objective by $30\%$ on both sides, so as to explore outside the currently observed boundaries. All perturbed objective values are clipped back to this extended interval before normalization and reward computation, which empirically provides a good trade-off between reward diversity and training stability.

\begin{table}[htbp]
    \centering
    \caption{Training configuration.}
    \label{tab:training_config}
    \resizebox{0.7\linewidth}{!}{
    \begin{tabular}{|p{0.3\linewidth}|p{0.7\linewidth}|}
    \hline
    Aspect & Setting \\ \hline
    $\lambda_{\mathrm{CQL}}$(offline Q-learning) & 0.1\\ \hline
    $\gamma$(offline Q-learning) & 0.99 \\ \hline
    Mantissa digits $n_{digit}$ & 15 \\ \hline
    Backbone & T5 family (\texttt{t5-small} if per-task training size $<1000$; otherwise \texttt{t5-base}) \\ \hline
    Max input / target length & $20\times D$ tokens (covers metadata + $x$; e.g., $D{=}50\Rightarrow 1000$, $D{=}60\Rightarrow 1200$) \,/\, 50 tokens \\ \hline
    Tokenizer & \texttt{T5Tokenizer} \\ \hline
    Optimizer & Adafactor, learning rate $1\times 10^{-3}$, weight decay $0$ \\ \hline
    Learning rate schedule & Constant-with-warmup, warmup ratio $0.06$ \\ \hline
    Batch size (per GPU) & \texttt{t5-small}: 24; \texttt{t5-base}: 12 \\ \hline
    Epochs & 65 (SFT) $+$ 5 (offline Q-learning) \\ \hline
    Precision & FP32 \\ \hline
    Hardware & NVIDIA GeForce RTX 4090 \\ \hline
    \end{tabular}}
\end{table}

\subsubsection{Comparison Experiments of Surrogate Models}
\begin{table}[htbp]
\caption{Surrogate baselines included in our comparison. Architectures and training hyperparameters strictly follow official papers unless stated otherwise.}
\label{tab:surrogate_baselines}
\footnotesize
\centering
    \resizebox{1\linewidth}{!}{
    \begin{tabular}{|p{0.2\linewidth}| p{0.8\linewidth}|}
\hline
Baseline & Model configuration \\
\hline
RBFN~\cite{CC-DDEA} &
Single-task Gaussian radial basis network per objective and per problem dimension; centers from $k$-means/$k$-means++ on training inputs; basis width $\sigma_j$ scaled by the mean distance to nearest neighbors; output weights by least squares; the number of centers selected by a small validation grid. \\
\hline
KAN (Kolmogorov--Arnold Networks)~\cite{KAN} &
Edge-level spline activations based on the Kolmogorov--Arnold representation: each ``weight'' is a learnable univariate B-spline. We adopt compact widths and shallow depth following \texttt{pykan} examples; optimize with MSE loss and Adam by default; optional sparsification (set $\lambda{=}0$ if targeting pure accuracy). During training, \texttt{speed()} disables the symbolic branch. In multi-objective settings, train one KAN per objective. Grid size/order/layer width are selected via a small validation sweep per official guidance. \\\hline
FTGP (ExTrEMO)~\cite{FTGP}\tablefootnote{\url{https://github.com/LiuJ-2023/ExTrEMO}} &
Factorized transfer Gaussian process trained at the granularity of (domain, instance, task, objective). For each objective, the target task is aided by source tasks from the same instance. Train a separate FTGP per objective. First fit an ExactGP for the target (constant mean; RBF kernel with length-scale constrained to $[0.05,10.0]$, initialized to $0.5$; Gaussian likelihood with noise constrained to $[10^{-4},5]$) by maximizing exact marginal likelihood (Adam, lr$=0.1$, 45 iterations). For each source--target pair, build a transfer GP with product kernel $k_x(x,x')\otimes k_{\mathrm{task}}(i,i')$, copying and freezing the target GP's mean, length-scale, and noise. Predictions fuse factorized experts multiplicatively. \\\hline
\end{tabular}
    }
\end{table}

Under a unified data split and compute budget, we compare the proposed meta-surrogate against three representative alternatives: single-task radial basis function networks (RBFN), KANs based on the Kolmogorov--Arnold representation, and the transfer multi-task GP FTGP in ExTrEMO (Table~\ref{tab:surrogate_baselines}). All models are trained/evaluated on the same train/validation/test partitions. Metrics are sMAE ($\downarrow$) and $R^2$ ($\uparrow$).

Overall, Tables~\ref{tab:tradition_resulta}--\ref{tab:tradition_resultb} show that Q-MetaSur attains the best sMAE on 10 out of 12 instance--objective pairs, with correspondingly top $R^2$; the remaining two pairs (Inst2/Obj1 and Inst3/Obj1) are led by RBFN. Notably, our model achieves zero sMAE with $R^2{=}1.0000$ on Inst2/Obj0, Inst3/Obj0, and Inst5/Obj0, indicating near-exact reconstruction for certain deterministic mappings. On Inst1, Inst4, and Inst6, Q-MetaSur is also markedly better than the three baselines (e.g., for Inst1/Obj0--1, sMAE is 0.0693/0.0655 and $R^2$ is 0.8316/0.8379; for Inst6/Obj0--1, sMAE is 0.0511/0.0528 and $R^2$ is 0.9101/0.9044). FTGP can be competitive in several settings (e.g., Inst5) but typically yields lower $R^2$ than Q-MetaSur overall. KAN exhibits more pronounced $R^2$ drops under limited data and highly rugged objectives, suggesting robustness challenges in small-sample, multi-task, multi-objective regression.

By instance type, the six instances cover common multi-objective landscapes: near-spherical (Inst1), mean-scale transformation (Inst2), Rosenbrock-like with narrow valleys and strong nonlinear coupling (Inst3), highly multimodal Rastrigin (Inst4), and Ackley (Inst5) and Griewank (Inst6). Q-MetaSur shows stable advantages in most types, especially in multimodal or weakly separable settings (Inst4, Inst6), where metadata conditioning and cross-task sharing substantially improve both error and $R^2$ over traditional surrogates. For Inst2/Obj1 and Inst3/Obj1---objectives with more localized structure---RBFN benefits from local basis functions within a small data regime, whereas the meta-surrogate trades off unified modeling for cross-task robustness. This suggests a practical avenue: retain a single meta-surrogate while adding a small expert ensemble at inference time to absorb local idiosyncrasies and further tighten the tail error. In sum, the proposed method improves accuracy and stability on the vast majority of instance--objective pairs while preserving the transferability of ``one model'' across tasks/objectives/landscapes.

\begin{table}[htbp]
\caption{Per-instance, per-objective sMAE ($\downarrow$); best across models in \textbf{bold}.}
\label{tab:tradition_resulta}
\footnotesize
\centering
\resizebox{0.6\linewidth}{!}{
\begin{tabular}{l cc cc cc cc}
\toprule
\multirow{2}{*}{Instance} & \multicolumn{2}{c}{RBFN} & \multicolumn{2}{c}{KAN} & \multicolumn{2}{c}{FTGP} & \multicolumn{2}{c}{Q-MetaSur} \\
         & Obj0 & Obj1 & Obj0 & Obj1 & Obj0 & Obj1 & Obj0 & Obj1 \\
\midrule
CEC19\_MTMOO-Inst1 & 0.2064 & 0.2026 & 0.4582 & 0.4357 & 0.2077 & 0.2023 & \textbf{0.0693} & \textbf{0.0655} \\
CEC19\_MTMOO-Inst2 & 0.2615 & \textbf{0.1083} & 1.8741 & 1.0453 & 0.2510 & 0.1494 & \textbf{0.0000} & 0.1529 \\
CEC19\_MTMOO-Inst3 & 0.2513 & \textbf{0.0944} & 0.2533 & 0.6687 & 0.2458 & 0.1532 & \textbf{0.0000} & 0.1587 \\
CEC19\_MTMOO-Inst4 & 0.1908 & 0.1860 & 0.4278 & 0.4232 & 0.1956 & 0.1980 & \textbf{0.0627} & \textbf{0.0610} \\
CEC19\_MTMOO-Inst5 & 0.1221 & 0.0959 & 0.1397 & 0.1824 & 0.0443 & 0.0797 & \textbf{0.0000} & \textbf{0.0623} \\
CEC19\_MTMOO-Inst6 & 0.2173 & 0.2223 & 0.2236 & 0.2400 & 0.2110 & 0.2144 & \textbf{0.0511} & \textbf{0.0528} \\
\bottomrule
\end{tabular}}
\end{table}

\begin{table}[htbp]
\caption{Per-instance, per-objective $R^2$ ($\uparrow$); best across models in \textbf{bold}.}
\label{tab:tradition_resultb}
\footnotesize
\centering
\resizebox{0.8\linewidth}{!}{
\begin{tabular}{l cc cc cc cc}
\toprule
\multirow{2}{*}{Instance} & \multicolumn{2}{c}{RBFN} & \multicolumn{2}{c}{KAN} & \multicolumn{2}{c}{FTGP} & \multicolumn{2}{c}{Q-MetaSur} \\
         & Obj0 & Obj1 & Obj0 & Obj1 & Obj0 & Obj1 & Obj0 & Obj1 \\
\midrule
CEC19\_MTMOO-Inst1 & -0.0751 & -0.1021 & -3.8994 & -3.7700 & -0.0611 & -0.0625 & \textbf{0.8316} & \textbf{0.8379} \\
CEC19\_MTMOO-Inst2 & -0.1994 & \textbf{0.4367} & -72.8734 & -34.7733 & -0.0567 & -0.0370 & \textbf{1.0000} & -0.0918 \\
CEC19\_MTMOO-Inst3 & -0.1242 & \textbf{0.5624} & -0.2909 & -14.0647 & -0.0411 & -0.0579 & \textbf{1.0000} & -0.1558 \\
CEC19\_MTMOO-Inst4 & 0.0248 & 0.0788 & -3.3217 & -3.3650 & 0.0106 & -0.0125 & \textbf{0.8630} & \textbf{0.8686} \\
CEC19\_MTMOO-Inst5 & 0.7146 & 0.6419 & 0.6364 & -0.1193 & 0.8468 & 0.7372 & \textbf{1.0000} & \textbf{0.8474} \\
CEC19\_MTMOO-Inst6 & -0.1431 & -0.1483 & -0.3208 & -0.4046 & -0.0504 & -0.0422 & \textbf{0.9101} & \textbf{0.9044} \\
\bottomrule
\end{tabular}}
\end{table}

\subsection{Emergent Capability (RQ2)}
We examine the \emph{emergent generalization} of Q-MetaSur, i.e., the ability to deliver reliable predictions outside the training distribution. We evaluate two orthogonal aspects. (i) \emph{Unseen-task generalization}: under fixed dimension and instance family, we generate entirely new task parameterizations via rotation and shift matrices consistent with \textsc{MetaBox}; these tasks never appear during training, and the model infers zero-shot. (ii) \emph{Unseen-dimension generalization}: with the instance family fixed, we increment the decision dimension beyond the training baseline and perform zero-shot inference without any additional training samples. The former probes structural extrapolation within-task families; the latter probes representational and length extrapolation across dimensions.

Training/testing are as follows. The six instances of the CEC2019 multi-task benchmark (each with 50 tasks) are used for offline training; each task contributes 200 samples. As a baseline, RBFN is fitted per task: in the unseen-task setting, we provide 20 training samples per new task; in the unseen-dimension setting, due to limited transfer across dimensions, we provide only 5 samples. Q-MetaSur (T5 backbone) is evaluated zero-shot in both settings. Metrics are sMAE ($\downarrow$) and $R^2$ ($\uparrow$).

\begin{table*}[htbp]
\centering
\footnotesize
\setlength{\tabcolsep}{3.2pt}
\caption{Unseen-task generalization on CEC2019 MTMOO: instance-level averages over 50 generated tasks per instance. Q-MetaSur is evaluated zero-shot; RBFN is trained per new task with 20 samples. Lower sMAE and higher $R^2$ indicate better accuracy. Boldface highlights the better method in each cell. Obj.~1/2 correspond to the first/second objective.}

\label{tab:inst-avg}
\begin{tabular}{lcccccccc}
\toprule
\multirow{2}{*}{Instance} &
\multicolumn{2}{c}{Q-MetaSur, Obj.~1 ($o_0$)} &
\multicolumn{2}{c}{Q-MetaSur, Obj.~2 ($o_1$)} &
\multicolumn{2}{c}{RBFN, Obj.~1 ($o_0$)} &
\multicolumn{2}{c}{RBFN, Obj.~2 ($o_1$)} \\
& sMAE $\downarrow$ & $R^2$ $\uparrow$
& sMAE $\downarrow$ & $R^2$ $\uparrow$
& sMAE $\downarrow$ & $R^2$ $\uparrow$
& sMAE $\downarrow$ & $R^2$ $\uparrow$ \\
\midrule
Inst1 & \textbf{0.0691} & \textbf{0.8386} & \textbf{0.0698} & \textbf{0.8259} & 0.2604 & -0.8757 & 0.2553 & -0.8401\\
Inst2 & \textbf{0.0000} & \textbf{1.0000} & 0.1470 & \textbf{-0.0310} & 0.3313 & -1.0005 & \textbf{0.1464} & -0.0389\\
Inst3 & \textbf{0.0000} & \textbf{1.0000} & 0.1556 & -0.1054 & 0.3224 & -0.8701 & \textbf{0.1338} & \textbf{0.1550}\\
Inst4 & \textbf{0.0645} & \textbf{0.8680} & \textbf{0.0642} & \textbf{0.8589} & 0.2564 & -0.6534 & 0.2484 & -0.6415\\
Inst5 & \textbf{0.0000} & \textbf{1.0000} & \textbf{0.0638} & \textbf{0.8467} & 0.2393 & -0.0418 & 0.1787 & -0.1658\\
Inst6 & \textbf{0.0544} & \textbf{0.9126} & \textbf{0.0551} & \textbf{0.9050} & 0.2763 & -0.8167 & 0.2723 & -0.8039\\
\bottomrule
\end{tabular}
\end{table*}

For unseen tasks, instance-averaged results are given in Table~\ref{tab:inst-avg}. Q-MetaSur achieves clear advantages on most instance--objective pairs: for Inst1, Inst4, and Inst6, both objectives exhibit the favorable combination of low error and high $R^2$, indicating that cross-task structural priors have been internalized. A representative pattern is the ``near-deterministic'' mapping on Obj0: Inst2/3/5 reach sMAE$=0.0000$ and $R^2{=}1.0000$, showing that when the objective exhibits clear global scaling and monotonicity, the meta-surrogate can reconstruct it nearly exactly under a unified representation. Two exceptions remain: on Inst2/Obj1 and Inst3/Obj1, RBFN is slightly better, likely due to sharper local curvature and narrow valleys where local bases benefit from bias under small samples. Even so, Q-MetaSur maintains usable sMAE and $R^2$ without per-task retraining---a crucial advantage in offline, compute- and data-constrained scenarios. Overall, zero-shot performance on unseen tasks suggests that the unified encoding of textual metadata and scientific-notation sequences enables the model to invoke learned family invariances (e.g., units, sign, magnitude, and coarse coupling patterns) on new tasks.

\begin{table*}[htbp]
\centering
\footnotesize
\setlength{\tabcolsep}{3.2pt}
\caption{Dimension-wise performance across all six CEC2019 MTMOO instances. Each entry is averaged over instances; within each instance, metrics are first averaged over its 50 generated tasks at the given dimension. Q-MetaSur is zero-shot; RBFN is retrained per new dimension with 5 samples. Lower sMAE and higher $R^2$ are better; boldface marks the better method per cell. Obj.~1/2 correspond to the first/second objective.}

\label{tab:overall_dim_avg}
\begin{tabular}{lcccccccc}
\toprule
\multirow{2}{*}{Dim.} &
\multicolumn{2}{c}{Q-MetaSur, Obj.~1} &
\multicolumn{2}{c}{Q-MetaSur, Obj.~2} &
\multicolumn{2}{c}{RBFN, Obj.~1} &
\multicolumn{2}{c}{RBFN, Obj.~2} \\
& sMAE $\downarrow$ & $R^2$ $\uparrow$
& sMAE $\downarrow$ & $R^2$ $\uparrow$
& sMAE $\downarrow$ & $R^2$ $\uparrow$
& sMAE $\downarrow$ & $R^2$ $\uparrow$ \\
\midrule
21 & \textbf{0.0322} & \textbf{0.9261} & \textbf{0.0980} & \textbf{0.5167} & 0.2653 & -0.4775 & 0.2053 & -0.2931\\
22 & \textbf{0.0341} & \textbf{0.9228} & \textbf{0.0969} & \textbf{0.5394} & 0.2498 & -0.3105 & 0.1976 & -0.2313\\
23 & \textbf{0.0402} & \textbf{0.9060} & \textbf{0.1113} & \textbf{0.4203} & 0.2526 & -0.2719 & 0.2057 & -0.3293\\
24 & \textbf{0.0492} & \textbf{0.8611} & \textbf{0.1175} & \textbf{0.3907} & 0.2569 & -0.4056 & 0.2090 & -0.3231\\
25 & \textbf{0.0480} & \textbf{0.8672} & \textbf{0.1201} & \textbf{0.3549} & 0.2607 & -0.3670 & 0.2113 & -0.2987\\
26 & \textbf{0.0573} & \textbf{0.8229} & 1.4415 & -1149 & 0.2534 & -0.2767 & \textbf{0.2063} & \textbf{-0.3176}\\
27 & \textbf{0.0595} & \textbf{0.8128} & \textbf{0.1374} & \textbf{0.2355} & 0.2649 & -0.4089 & 0.2122 & -0.3314\\
28 & \textbf{0.0676} & \textbf{0.7634} & \textbf{0.1462} & \textbf{-0.0578} & 0.2548 & -0.2844 & 0.2023 & -0.2368\\
\bottomrule
\end{tabular}
\end{table*}

For unseen dimensions, Table~\ref{tab:overall_dim_avg} (averaged over 50 tasks and all six instances) shows that as the dimension increments from the training baseline towards 21--28, Q-MetaSur maintains clear advantages: most instance--objective pairs yield sMAE in the low range $[0.05,\,0.14]$, with $R^2$ mostly positive in $[0.47,\,0.90]$, and generally better than RBFN trained with only 5 samples per new dimension. This reveals two points. First, the model learns a hierarchical decoding of digits--exponent--sign: it aligns magnitude and sign before refining mantissas, so longer inputs due to moderate dimensional increments do not immediately destabilize attention. Second, the ``dimension'' tag in metadata supplies a conditional prior on sequence length and structure, placing 21--28D extrapolation within a tractable ``increment band.'' 



\subsection{Performance Evaluation on MTMOO (RQ3)}
We evaluate optimization gains on the CEC2019 MTMOO benchmark: six instances, each with 50 tasks. To ensure comparability, all methods use the same budget of $200$ \emph{true} function evaluations (FEs). For the \emph{REAL} baselines (true-fitness backbones), the run terminates once the cumulative FEs reach the budget. For offline surrogates, training data are generated once via Latin Hypercube Sampling (LHS) \emph{before} the algorithm runs, and then used to train the surrogate; during optimization, no further true evaluations are performed. The entire evolution uses surrogate predictions in place of fitness, and only the final best individual is evaluated once by the true objectives for reporting. Each problem is run 20 times; we report the mean and standard deviation, and use the Wilcoxon \emph{signed-rank} test ($\alpha{=}0.05$) for paired comparisons. In the tables, ``\(+\)'', ``\(\approx\)'', and ``\(-\)'' denote significantly better than LLM, no significant difference, and significantly worse than LLM, respectively.

\textit{Metrics and aggregation.} Within each task, performance is measured by the inverted generational distance (IGD) and further aggregated into an instance-level mean standardized score (MSS) for algorithm ranking. 

For a test instance with $K$ tasks, let $I_i$ be the IGD of task $T_i$, and compute the per-task mean $\mu_i$ and standard deviation $\sigma_i$ using \emph{all} algorithms and all runs. The instance-level MSS is
\begin{equation}
\label{eq:mss_rq3}
\mathrm{MSS}\;=\;\frac{1}{K}\sum_{i=1}^{K}\frac{I_i-\mu_i}{\sigma_i},
\end{equation}
where smaller values (especially significantly negative) indicate better and more stable overall IGD on that instance. 
We integrate the proposed Q\text{-}MetaSur into two backbones, MO-MaTDE and MTEA-DCK. Similarly, we form offline DDEA variants by replacing the Q\text{-}MetaSur with RBFN, FTGP, or KAN. MO-MaTDE uses multiple subpopulations and adaptive transfer selection, identifying helper tasks by KL similarity and cumulative transfer returns, and sharing knowledge via crossover-based transfers. MTEA-DCK partitions the population into exploration-oriented ``winners'' and convergence-oriented ``losers'', and combines DKT/CKT (on particle states) with differential evolution and competitive swarm optimization to decouple transfer from convergence.

\begin{table}[t]
\centering
\footnotesize
\caption{Mean standardized score (MSS; lower is better) aggregated over $K{=}50$ tasks for each instance--algorithm pair under a strict $200$-FE budget. For non-LLM columns, superscripts indicate significance against the LLM (Q\text{-}MetaSur) at $\alpha{=}0.05$ using the Wilcoxon signed-rank test for paired samples: $(+)$ LLM significantly better; $(\approx)$ no significant difference; $(-)$ LLM significantly worse.}
\label{tab:mss-one-big}
\begin{tabular}{lccccc}
\toprule
Instance\_Algo & \multicolumn{1}{c}{Q\text{-}MetaSur} & \multicolumn{1}{c}{REAL} & \multicolumn{1}{c}{RBFN} & \multicolumn{1}{c}{FTGP} & \multicolumn{1}{c}{KAN} \\
\midrule
ins1\_MO-MaTDE & \textbf{-0.777 $\pm$ 0.065} & -0.153 $\pm$ 0.041{(+)} & -0.645 $\pm$ 0.064{(+)} & 0.852 $\pm$ 0.105{(+)} & 1.031 $\pm$ 0.265{(+)} \\
ins1\_MTEA-DCK & \textbf{-0.715 $\pm$ 0.102} & -0.154 $\pm$ 0.042{(+)} & -0.587 $\pm$ 0.176{($\approx$)} & 0.205 $\pm$ 0.083{(+)} & 0.942 $\pm$ 0.189{(+)} \\
ins2\_MO-MaTDE & \textbf{-1.035 $\pm$ 0.052} & -0.566 $\pm$ 0.071{(+)} & 0.732 $\pm$ 0.037{(+)} & 0.206 $\pm$ 0.087{(+)} & 0.365 $\pm$ 0.051{(+)} \\
ins2\_MTEA-DCK & \textbf{-1.082 $\pm$ 0.030} & -0.581 $\pm$ 0.031{(+)} & 1.946 $\pm$ 0.039{(+)} & -0.229 $\pm$ 0.106{(+)} & 0.242 $\pm$ 0.076{(+)} \\
ins3\_MO-MaTDE & \textbf{-0.936 $\pm$ 0.024} & -0.510 $\pm$ 0.048{(+)} & 1.568 $\pm$ 0.095{(+)} & 0.593 $\pm$ 0.079{(+)} & 0.633 $\pm$ 0.031{(+)} \\
ins3\_MTEA-DCK & \textbf{-1.230 $\pm$ 0.034} & -0.473 $\pm$ 0.104{(+)} & 0.549 $\pm$ 0.073{(+)} & -0.293 $\pm$ 0.040{(+)} & 0.099 $\pm$ 0.079{(+)} \\
ins4\_MO-MaTDE & -0.756 $\pm$ 0.043 & \textbf{-0.763 $\pm$ 0.047}($\approx$) & -0.148 $\pm$ 0.112{(+)} & 0.646 $\pm$ 0.065{(+)} & 1.171 $\pm$ 0.175{(+)} \\
ins4\_MTEA-DCK & \textbf{-0.771 $\pm$ 0.070} & -0.734 $\pm$ 0.089{($\approx$)} & -0.342 $\pm$ 0.091{(+)} & 0.538 $\pm$ 0.108{(+)} & 1.159 $\pm$ 0.149{(+)} \\
ins5\_MO-MaTDE & \textbf{-0.565 $\pm$ 0.027} & -0.528 $\pm$ 0.020{($\approx$)} & -0.342 $\pm$ 0.024{(+)} & -0.509 $\pm$ 0.010{(+)} & 1.755 $\pm$ 0.060{(+)} \\
ins5\_MTEA-DCK & \textbf{-0.670 $\pm$ 0.024} & -0.535 $\pm$ 0.012{(+)} & -0.003 $\pm$ 0.010{(+)} & -0.625 $\pm$ 0.022{(+)} & 2.022 $\pm$ 0.061{(+)} \\
ins6\_MO-MaTDE & \textbf{-1.028 $\pm$ 0.030} & -0.675 $\pm$ 0.047{(+)} & 1.586 $\pm$ 0.041{(+)} & 0.270 $\pm$ 0.092{(+)} & 0.561 $\pm$ 0.121{(+)} \\
ins6\_MTEA-DCK & \textbf{-1.034 $\pm$ 0.018} & -0.633 $\pm$ 0.052{(+)} & 0.748 $\pm$ 0.032{(+)} & -0.405 $\pm$ 0.063{(+)} & 0.610 $\pm$ 0.080{(+)} \\
\bottomrule
\end{tabular}
\end{table}

\textit{Overall results and significance.} Table~\ref{tab:mss-one-big} reports instance-level MSS on both backbones. Q\mbox{-}MetaSur attains the best MSS on \(11\) out of \(12\) instance--backbone pairs and is significantly better than the REAL backbones in most instances (``\(+\)'' in the REAL column); \texttt{ins4\_MO\mbox{-}MaTDE} is a statistical tie (``\(\approx\)''). For example, with MTEA\mbox{-}DCK, the MSS of \texttt{ins2} and \texttt{ins6} are \(-1.082\pm 0.030\) and \(-1.034\pm 0.018\), significantly better than the corresponding REAL values \(-0.581\pm 0.031\) and \(-0.633\pm 0.052\). Compared with other surrogates, RBFN and KAN tend to yield positive MSS (weaker than average) on most instances; FTGP approaches the LLM on a few cases but lags overall.

\textit{Analysis and discussion.} Under a strict 200-FE budget and an offline constraint of ``no additional FEs'' during optimization, the meta-surrogate reliably infers objective values across tasks, thereby providing more accurate selection signals (e.g., multi-objective ranking and crowding distance), which drive more effective approximation to the PF in limited iterations. The effect is clearest on \texttt{ins2}/\texttt{ins3}/\texttt{ins6}, where scaling changes and nonlinear coupling are present. On highly multimodal \texttt{ins4}, the tie with REAL suggests that in a few landscapes, the backbone with limited true evaluations can extract a small advantage; however, no case shows the LLM being significantly \emph{worse}. Importantly, \emph{Q\mbox{-}MetaSur is never significantly inferior to REAL} across instance--backbone pairs, providing statistical evidence for replacing true evaluations with the surrogate in cost-sensitive workflows. Overall, Q\mbox{-}MetaSur delivers stable, significant, and reproducible MSS improvements on MO\mbox{-}MaTOP for both multitask backbones under stringent FE budgets and offline data constraints, enabling a single-model surrogate to cover multi tasks/objectives with statistically significant gains and variance stability.

\begin{figure}[htbp]
    \centering
    \subfloat[\label{fig:ins1}]{
        \includegraphics[height=0.32\textwidth]{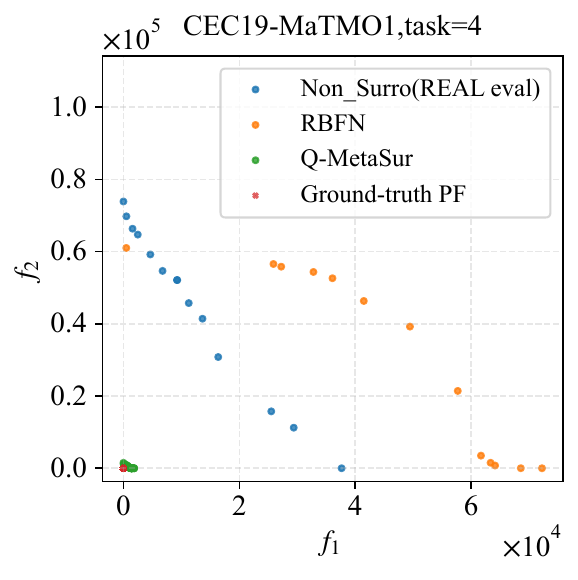}
    }
    \subfloat[\label{fig:ins2}]{
        \includegraphics[height=0.32\textwidth]{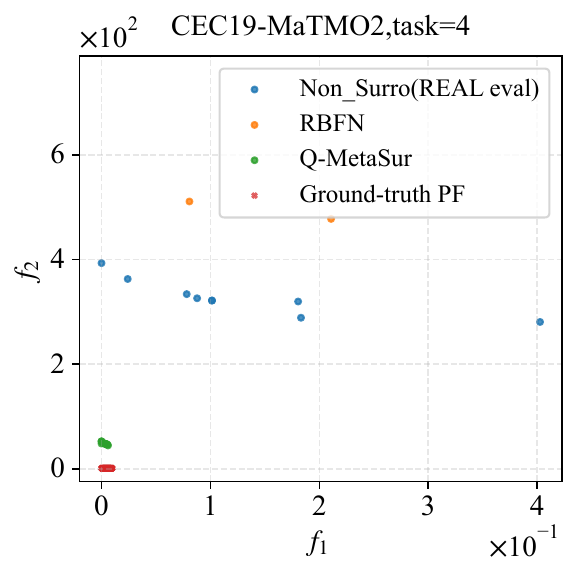}
    }
    \subfloat[\label{fig:ins3}]{
        \includegraphics[height=0.32\textwidth]{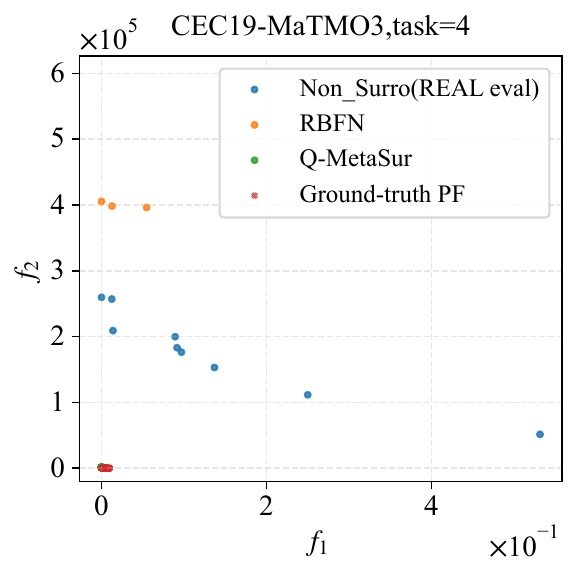}
    }\\
    \subfloat[\label{fig:ins4}]{
        \includegraphics[height=0.32\textwidth]{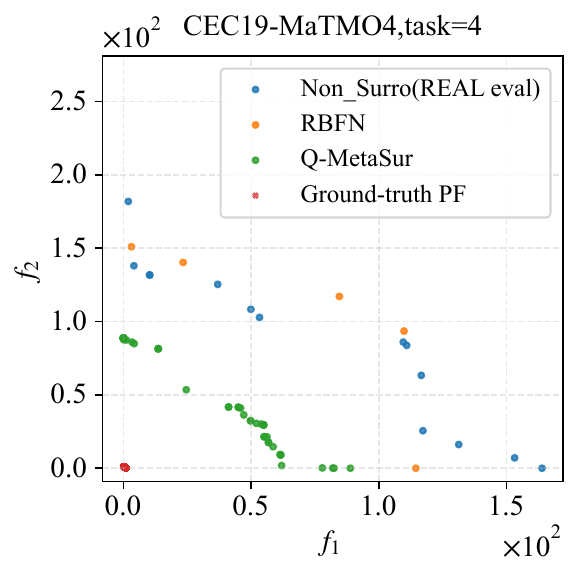}
    }
    \subfloat[\label{fig:ins5}]{
        \includegraphics[height=0.32\textwidth]{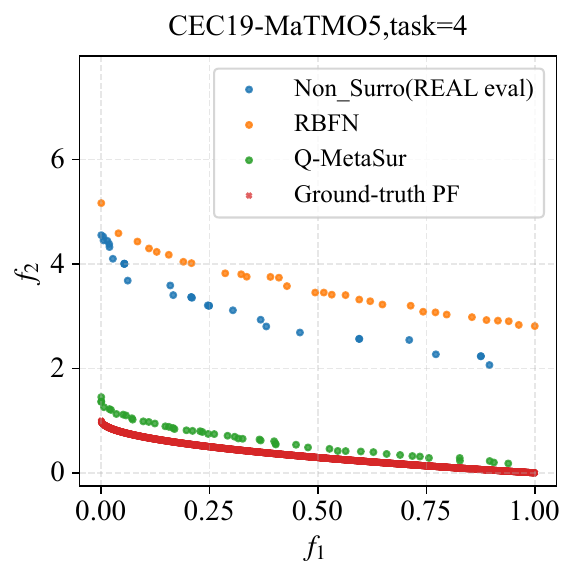}
    }
    \subfloat[\label{fig:ins6}]{
        \includegraphics[height=0.32\textwidth]{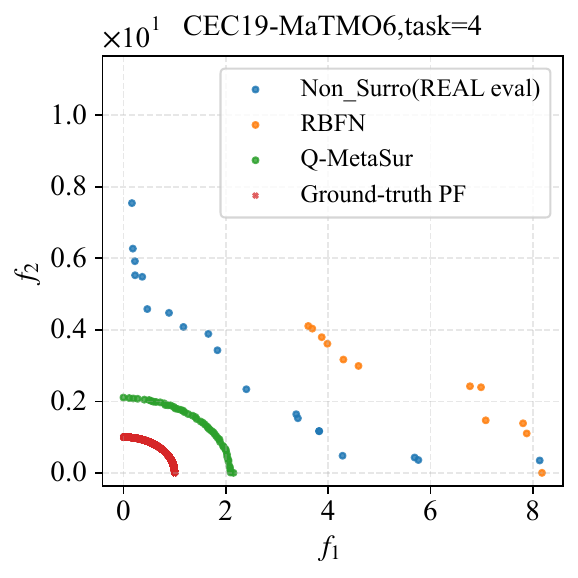}
    }
    \caption{Pareto fronts on a representative task from the six CEC2019 MTMOO instances under the MTEA-DCK backbone with different surrogates (Q\text{-}MetaSur, RBFN) and the REAL baseline. Each panel corresponds to one instance, and the reference ground-truth PF is also shown for comparison.}
    \label{fig:pf-mteadck}
\end{figure}

To complement the MSS statistics with a qualitative view, we further visualize the Pareto fronts obtained on a representative task shared by the six CEC2019 MTMOO instances when MTEA-DCK is coupled with different surrogates. For each instance, we plot the final non-dominated solutions returned by the Q\text{-}MetaSur-, and RBFN-based variants, together with the PF obtained by the REAL backbone and the reference ground-truth PF. As shown in Fig.~\ref{fig:pf-mteadck}, the PF produced by the Q\text{-}MetaSur-accelerated MTEA-DCK consistently dominates those of the other surrogates across all six instances. On instances~1 and~3 in particular, the PF obtained with Q\text{-}MetaSur almost coincides with the reference ground-truth PF, indicating that the learned LLM-based meta-surrogate can drive the backbone optimizer to nearly best-possible solutions under the strict 200-FE budget.

\subsection{In-Depth Analysis}

\subsubsection{Ablation Study}
\begin{table*}[htbp]
\centering
\footnotesize
\caption{Comparison by \textbf{sMAE (lower is better)}. Each cell is the instance average over $50$ tasks $\times$ $2$ objectives; ``Avg'' is the arithmetic mean over the six instances.}
\label{tab:smae_only}
\setlength{\tabcolsep}{4pt}
\begin{adjustbox}{width=0.8\textwidth}
\begin{tabularx}{0.8\textwidth}{@{}*{8}{>{\centering\arraybackslash}X}@{}}
\toprule
Instance &
\shortstack{SFT\\\footnotesize(PWCE)} &
\shortstack{SFT\\\footnotesize(CE)} &
\shortstack{$\mathrm{Q\mbox{-}MetaSur}$\\\footnotesize(Full)} &
\shortstack{$\mathrm{Q\mbox{-}MetaSur}$\\\footnotesize(Greedy)} &
\shortstack{$\mathrm{Q\mbox{-}MetaSur}$\\\footnotesize(w/o-AuxCE)} &
\shortstack{$\mathrm{Q\mbox{-}MetaSur}$\\\footnotesize(w/o-CQL)} &
\shortstack{$\mathrm{Q\mbox{-}MetaSur}$\\\footnotesize(w/o-PWCE)} \\
\midrule
Inst1   & 0.0815 & 0.0837 & 0.0674 & 0.0675 & \textbf{0.0670} & \textbf{0.0670} & 0.0692 \\
Inst2   & 0.1046 & 0.1123 & \textbf{0.0764} & \textbf{0.0764} & 0.0770 & 0.0797 & 0.0824 \\
Inst3   & 0.1133 & 0.1232 & 0.0794 & 0.0795 & \textbf{0.0787} & 0.0836 & 0.0864 \\
Inst4   & 0.0751 & 0.0760 & \textbf{0.0619} & 0.0620 & 0.0773 & 0.2062 & 0.0630 \\
Inst5   & 0.0408 & 0.0449 & \textbf{0.0310} & 0.0311 & 0.0315 & 0.0342 & 0.0342 \\
Inst6   & 0.0615 & 0.0643 & 0.0520 & \textbf{0.0518} & 0.0527 & 0.0526 & 0.0541 \\
\bottomrule
\end{tabularx}
\end{adjustbox}
\end{table*}

\begin{table*}[htbp]
\centering
\footnotesize
\caption{Comparison by \textbf{$R^2$ (higher is better)}. Each cell is the instance average over $50$ tasks $\times$ $2$ objectives; ``Avg'' is the arithmetic mean over the six instances. Negative values indicate collapse relative to the mean predictor. Column abbreviations follow Table~\ref{tab:smae_only}.}
\label{tab:r2_only}
\small
\setlength{\tabcolsep}{4pt}
\begin{adjustbox}{width=0.8\textwidth}
\begin{tabularx}{0.8\textwidth}{@{}*{8}{>{\centering\arraybackslash}X}@{}}
\toprule
Instance &
\shortstack{SFT\\\footnotesize(PWCE)} &
\shortstack{SFT\\\footnotesize(CE)} &
\shortstack{$\mathrm{Q\mbox{-}MetaSur}$\\\footnotesize(Full)} &
\shortstack{$\mathrm{Q\mbox{-}MetaSur}$\\\footnotesize(Greedy)} &
\shortstack{$\mathrm{Q\mbox{-}MetaSur}$\\\footnotesize(w/o-AuxCE)} &
\shortstack{$\mathrm{Q\mbox{-}MetaSur}$\\\footnotesize(w/o-CQL)} &
\shortstack{$\mathrm{Q\mbox{-}MetaSur}$\\\footnotesize(w/o-PWCE)} \\
\midrule
Inst1   & 0.8352 & 0.8252 & 0.8347 & 0.8339 & \textbf{0.8369} & 0.8354 & 0.8252 \\
Inst2   & 0.4492 & 0.3513 & \textbf{0.4548} & 0.4525 & 0.4483 & 0.4300 & 0.3515 \\
Inst3   & 0.4150 & 0.2868 & 0.4221 & 0.4193 & \textbf{0.4285} & 0.4018 & 0.2945 \\
Inst4   & 0.8639 & 0.8615 & \textbf{0.8658} & 0.8651 & -11.3169 & -101.7811 & 0.8610 \\
Inst5   & 0.9232 & 0.9060 & \textbf{0.9237} & 0.9236 & 0.9218 & 0.9046 & 0.9067 \\
Inst6   & 0.9082 & 0.9025 & 0.9072 & \textbf{0.9083} & 0.9053 & 0.9048 & 0.9026 \\
\bottomrule
\end{tabularx}
\end{adjustbox}
\end{table*}

We conduct ablations along three axes---supervised losses, offline RL regularizers, and inference decoding---to isolate the marginal contributions to accuracy and stability. All variants follow the same data splits and reporting protocol as Tables~\ref{tab:smae_only}--\ref{tab:r2_only}, and metrics are instance-averaged sMAE (lower is better) and $R^2$ (higher is better) over $50$ tasks and $2$ objectives per instance. Variants:
\begin{itemize}
  \item $\mathrm{SFT}_{\mathrm{(PWCE)}}$: supervised fine-tuning with the proposed priority-weighted cross-entropy (PWCE).
  \item $\mathrm{SFT}_{\mathrm{(CE)}}$: supervised fine-tuning with standard LM cross-entropy.
  \item $\text{Q-MetaSur}_{\mathrm{(full)}}$: initialize with $\mathrm{SFT}_{\mathrm{PWCE}}$, then offline Q-learning fine-tuning with auxiliary LLM-CE and CQL regularization, using value-guided decoding at inference.
  \item $\text{Q-MetaSur}_{\mathrm{(Greedy)}}$: same training as Full but greedy decoding at inference instead of value guidance.
  \item $\text{Q-MetaSur}_{\mathrm{(w/o-AuxCE)}}$: remove the auxiliary $\mathcal{L}_{\mathrm{NLL}}$ term during offline RL; otherwise identical to Full.
  \item $\text{Q-MetaSur}_{\mathrm{(w/o-CQL)}}$: remove the conservative CQL regularizer during offline Q-learning; otherwise identical to Full.
  \item $\text{Q-MetaSur}_{\mathrm{(w/o-PWCE)}}$: replace PWCE with standard CE in SFT; offline RL identical to Full.
\end{itemize}

On averages across instances, the full $\mathrm{Q\mbox{-}MetaSur}$ achieves the best error and explanatory power: its sMAE is $0.0613$, about $22.8\%$ lower than $\mathrm{SFT}_{\mathrm{PWCE}}$ ($0.0795$); its $R^2$ is $0.7347$, slightly higher than $\mathrm{SFT}_{\mathrm{PWCE}}$ ($0.7325$). This indicates that under a unified text-based numeric representation and scientific-notation reward shaping, the \emph{SFT + offline RL} cascade not only reduces mean error substantially but also maintains robust gains in explained variance. Instance-wise, $\mathrm{Q\mbox{-}MetaSur}$ dominates or ties on sMAE and $R^2$ for Inst2--Inst5, while being close to the Greedy variant on Inst1/Inst6. This suggests that value guidance acts chiefly as a \emph{safety rail} against worst-case degradation rather than improving all-instance averages.

Loss design in SFT \emph{pre-shapes} downstream RL. Compared with $\mathrm{SFT}_{\mathrm{CE}}$, $\mathrm{SFT}_{\mathrm{PWCE}}$ achieves lower sMAE and higher $R^2$ on \emph{all} six instances (cf. Tables~\ref{tab:smae_only}--\ref{tab:r2_only}), indicating that differential weighting on sign, exponent, and leading mantissa better aligns sequence likelihood with numerical error. Propagating this prior into RL, $\text{Q-MetaSur}_{\mathrm{(w/o-PWCE)}}$ shows higher sMAE and lower $R^2$ than Full, evidencing PWCE's positive impact on \emph{consistency} and a smoother \emph{optimization path}. While RL can compensate imperfect SFT to some extent, it cannot fully erase initial alignment gaps.

The two offline RL regularizers are crucial for stability. Removing CQL ($\text{Q-MetaSur}_{\mathrm{(w/o-CQL)}}$) leads to catastrophic failure on Inst4: sMAE rises to $0.2062$ and $R^2$ collapses to $-101.78$, well below the mean predictor. This matches known over-optimism on \emph{unseen actions} in offline RL and confirms that vocabulary-level conservatism is indispensable for curbing inflated $\widetilde{Q}$ on out-of-distribution tokens. Dropping AuxCE ($\text{Q-MetaSur}_{\mathrm{(w/o-AuxCE)}}$) yields marginal gains or ties on a few instances (Inst1, Inst3) but severe $R^2$ collapse on Inst4 ($-11.32$), indicating that under heavy-tailed or multimodal targets, AuxCE maintains alignment between numeric fragments and LM semantics, preventing value learning from overfitting near local optima.

For decoding, value guidance provides little average gain over greedy but improves robustness under adversarial instances. On Inst4, Full maintains positive $R^2$ with low sMAE, whereas Greedy---although close to Full on benign cases like Inst6---lacks the \emph{re-normalization} effect and synergy with CQL, increasing risk on difficult cases. Viewed together, value-guided decoding extends the training-time conservative regularization into inference, with benefits concentrated in \emph{worst-case} robustness.

\subsubsection{Transfer Analysis at the Surrogate Level}
To quantify \emph{positive transfer} from multi-task learning (MTL) relative to single-task learning (STL), we compare sMAE (lower is better) and $R^2$ (higher is better; \texttt{scikit-learn} definition with range $(-\infty,1]$) on each ``instance--task--objective'' pair. We report two complementary indicators: absolute difference $\Delta$ and relative ratio $\mathrm{TR}$. For errors, $\Delta_{\mathrm{sMAE}}=\mathrm{sMAE}_{\mathrm{STL}}-\mathrm{sMAE}_{\mathrm{MTL}}$ and $\mathrm{TR}_{\mathrm{sMAE}}=\mathrm{sMAE}_{\mathrm{MTL}}/\mathrm{sMAE}_{\mathrm{STL}}$. For gains, $\Delta_{R^2}=R^2_{\mathrm{MTL}}-R^2_{\mathrm{STL}}$ and $\mathrm{TR}_{R^2}=(1-R^2_{\mathrm{MTL}})/(1-R^2_{\mathrm{STL}})$ (ratio of unexplained variance). Hence, $\Delta>0$ and $\mathrm{TR}<1$ indicate MTL superiority, and $\mathrm{TR}_{R^2}$ remains interpretable even when the baseline collapses ($R^2\!\ll\!0$).

\begin{table}[htbp]
\centering
\footnotesize
\caption{Instance-level surrogate transfer: arithmetic mean over $50$ tasks $\times$ $2$ objectives ($100$ pairs) per instance.}
\label{tab:transfer_resulta}
\resizebox{0.5\linewidth}{!}{
\begin{tabular}{lcccc}
\toprule
Instance & $\overline{\Delta}_{\mathrm{sMAE}}$ & $\overline{\mathrm{TR}}_{\mathrm{sMAE}}$ & $\overline{\Delta}_{R^2}$ & $\overline{\mathrm{TR}}_{R^2}$ \\
\midrule
Inst1 &  0.0234 & 0.8623 & 0.3727 & 0.5443 \\
Inst2 & 0.0128 & 0.9780 & 0.6883 & 0.7698 \\
Inst3 & 0.0156 & 0.9232 & 0.0873 & 0.8204 \\
Inst4 & 0.0877 & 0.5879 & 12.6259 & 0.2817 \\
Inst5 & 0.0253 & 0.6484 & 0.2702 & 0.4681 \\
Inst6 & 0.9815 & 0.7811 & 18237.5118 & 0.4993 \\
\bottomrule
\end{tabular}
}
\end{table}

Results in Table~\ref{tab:transfer_resulta} show broad and stable positive transfer across instances, tasks, and objectives: (i) error reduction is widespread ($\mathrm{TR}_{\mathrm{sMAE}}<1$), e.g., Inst4 and Inst5 reduce sMAE by about $41\%$ and $35\%$, while Inst2 remains a near tie with a small positive $\overline{\Delta}_{\mathrm{sMAE}}$; (ii) $\mathrm{TR}_{R^2}\!\approx\!0.50\pm0.05$ on Inst1/5/6, indicating $\sim\!50\%$ lower unexplained variance, with a strong correction on Inst4 ($\mathrm{TR}_{R^2}=0.2817$). When STL catastrophically extrapolates ($R^2\!\ll 0$), $\Delta_{R^2}$ can be extremely large (e.g., Inst6), which is not an anomaly but a direct reflection of MTL rescuing a failed baseline; in such cases, $\mathrm{TR}_{R^2}$ offers a more stable sense of improvement. Overall, the complementary use of $\Delta$ and $\mathrm{TR}$ avoids metric bias under strong/failed baselines and highlights the essence of shared representation in variance reduction and extrapolation robustness.

\section{Application Study}
\label{sec:app_experiments}

Sensor coverage optimization~\cite{MOCAP} aims to minimize the uncovered area within a given region in a cost-effective manner. It can be formulated as a bi-objective problem:
\begin{equation}
\begin{aligned}
\text{minimize}\quad 
& f_1(\mathbf{x})
= 1 -
\frac{
\mathcal{A}\!\left(
\mathcal{R}\ \cap\ \displaystyle\bigcup_{s=1}^{S}\,\pi r_s^{2}(u_s, v_s)
\right)
}{
\mathcal{A}(\mathcal{R})
},\\[0.2em]
& f_2(\mathbf{x})
= \displaystyle\sum_{s=1}^{S}\bigl(1 + 10\,r_s^{2}\bigr),\\[0.35em]
\text{subject to}\quad
& -1 < u_s < 1,\quad s=1,\ldots,S,\\
& -1 < v_s < 1,\quad s=1,\ldots,S,\\
& 0.1 < r_s < 0.25,\quad s=1,\ldots,S.
\end{aligned}
\tag{27}
\end{equation}

Here, $\mathbf{x}=(u_1,v_1,r_1,\ldots,u_S,v_S,r_S)\in\mathbb{R}^{3S}$ is the decision vector; $(u_s,v_s)$ and $r_s$ denote the center location and coverage radius of the $s$-th sensor, respectively; $\mathcal{A}(\cdot)$ is the area operator and $\mathcal{R}$ is the region to be covered (we take $\mathcal{R}=[-1,1]\times[-1,1]$ in our experiments).

Because the optimal number of sensors is unknown and the number of variables grows linearly with $S$, the problem can be viewed as a heterogeneous multi-task multi-objective setting. We consider $30$ different values of $S$, yielding $30$ optimization tasks with dimensions $\{6,9,\ldots,93\}$. Each task is allotted at most $500$ function evaluations (FEs), for a total budget of $15{,}000$ FEs; each algorithm is executed for $25$ independent runs. The results are summarized in Table~\autoref{tab:mss-mocap}, where Q\mbox{-}MetaSur achieves the best overall performance.

\begin{table*}[t]
\centering
\caption{MSS (lower is better) for each algorithm. Per row, the lowest mean is bolded. For non-LLM columns, superscripts indicate significance vs.\ LLM using the Wilcoxon signed-rank test at $\alpha=0.05$: $(+)$ LLM significantly better; $(-)$ LLM significantly worse; $(\approx)$ no significant difference.}
\label{tab:mss-mocap}
\resizebox{0.8\textwidth}{!}{%
\begin{tabular}{lccccc}
\toprule
Algo & \multicolumn{1}{c}{LLM(Q-MetaSur)} & \multicolumn{1}{c}{REAL} & \multicolumn{1}{c}{RBFN} & \multicolumn{1}{c}{FTGP} & \multicolumn{1}{c}{KAN} \\
\midrule
MOCAP\_MO-MaTDE & \textbf{-0.513 $\pm$ 0.066} & -0.001 $\pm$ 0.030$^{(+)}$ & 0.233 $\pm$ 0.052$^{(+)}$ & -0.088 $\pm$ 0.111$^{(+)}$ & 0.856 $\pm$ 0.092$^{(+)}$ \\
MOCAP\_MTEA-DCK & \textbf{-0.841 $\pm$ 0.148} & -0.068 $\pm$ 0.040$^{(+)}$ & 0.093 $\pm$ 0.211$^{(+)}$ & -0.043 $\pm$ 0.076$^{(+)}$ & 0.372 $\pm$ 0.268$^{(+)}$ \\
\bottomrule
\end{tabular}}
\end{table*}

\section{Conclusion and Outlook}\label{sec:conclusion}
This paper proposes treating a large language model (LLM) as a \emph{meta-surrogate} for \emph{multi-objective, multi-task} settings. By unifying textualized numerical representations---scientific-notation encoding (SNE) for numbers together with a templated metadata schema---we condition the surrogate on task--objective context. Building on \emph{offline} implicit Q-learning (expectile value learning, and vocabulary-level conservative regularization), we directly align sequence generation with continuous numerical error. Combined with priority-weighted cross-entropy (PWCE) and value-guided decoding, the proposed framework forms a closed loop at both the \emph{surrogate} and \emph{evolutionary} levels: the former shares cross-task knowledge within a single model and exhibits emergent generalization to unseen tasks and to moderate dimensionality increments (e.g., 21--28 dimensions); the latter, under a stringent function-evaluation (FE) budget and in an offline setting with no additional evaluations, serves as a drop-in \emph{fitness oracle} for MTMOO and consistently improves the convergence--diversity trade-off, yielding statistically significant and robust overall gains. More importantly, the experiments support a methodological conclusion: when representation, loss, value regularization, and decoding are designed in a \emph{support-set--centric} and consistent manner, an LLM can reliably carry offline multi-objective, multi-task regression \emph{without} problem-specific feature engineering, while maintaining engineering-grade scalability.

\textbf{Limitations.} The current approach has three main limitations: (i) \emph{Sensitivity near distribution boundaries.} When decision dimensionality moves beyond the training ``increment band'' (e.g., toward very high dimensions) or when task parameterization requires strong extrapolation, occasional stability degradation can occur, calling for few\mbox{-}shot correction to regain fast convergence; (ii) \emph{Dependence on rewards and metadata.} The fidelity of the offline reward design and the quality of metadata directly bound the ceiling of value learning and conditional modeling; insufficient task fingerprints or scale/unit encodings may diminish cross-task transfer gains; (iii) \emph{Inference and resource overhead.} Long numerical sequences (with SNE fragments) over a large vocabulary introduce latency and memory overheads; although pruning/caching help, there remains a gap to real-time performance.

\textbf{Future Work.} Building on these observations, we outline three complementary directions: (i) \emph{Adaptive calibration with few ground truths.} Introduce ultra-low-cost active calibration (few\mbox{-}shot true evaluations) and consistency regularization on top of the offline pipeline to close the loop from ``zero-shot prior'' to ``few-shot refinement''; (ii) \emph{Structured decoding and uncertainty enhancement.} Incorporate differentiable numerical-structure decoders (e.g., dedicated exponent--mantissa heads, range-alignment constraints) and uncertainty quantification with distribution calibration and monotonicity-preserving, coverage-guaranteed intervals (e.g., conformal prediction and temperature scaling), making the surrogate both accurate and \emph{well-calibrated}; (iii) \emph{Scalable expertization and heterogeneous transfer.} While preserving a single interface, introduce lightweight experts/adapters (MoE/task adapters) and multi-fidelity, multi-modal data fusion (textual fingerprints, landscape fingerprints, simulator logs) to better capture locally high-curvature or multi-modal terrains and further mitigate negative transfer under extreme heterogeneity.

Overall, this work provides a \emph{unified and scalable} surrogate-modeling route for offline multi-objective, multi-task optimization: using an LLM as the central carrier and coupling cross-task knowledge, numerical accuracy, and evolutionary selection via support-set--prioritized value learning and conservative regularization. We believe that, together with the above extensions, this line of research can continue to push the practical boundary of offline data-driven optimization toward larger scale, stronger heterogeneity, and harsher evaluation budgets.


\bibliography{QMetaSur_main}

\end{document}